\documentclass[11pt]{article}

\usepackage[preprint]{acl}

\usepackage{times}
\usepackage{latexsym}

\usepackage[T1]{fontenc}

\usepackage[utf8]{inputenc}

\usepackage{microtype}

\usepackage{inconsolata}

\usepackage{graphicx}

\usepackage{amssymb}
\usepackage{booktabs}
\usepackage{subcaption}
\usepackage{multirow}
\usepackage{wrapfig} 
\usepackage{colortbl}
\usepackage{enumitem}
\usepackage[most]{tcolorbox}
\usepackage{adjustbox}
\usepackage{cleveref}

\definecolor{Highlight}{HTML}{39b54a}  
\definecolor{Cerulean}{HTML}{00A2E3}  

\newcommand{\PAR}[1]{\noindent {\bf #1.~}} 

\usepackage[whole]{bxcjkjatype}

\usepackage{mdframed}
\definecolor{lightgray}{gray}{0.95}

\title{MangaVQA and MangaLMM: A Benchmark and Specialized Model for Multimodal Manga Understanding}

\author{%
Jeonghun Baek\thanks{Equal contribution.} \quad
Kazuki Egashira\footnotemark[1] \quad
Shota Onohara\footnotemark[1] \quad
Atsuyuki Miyai\footnotemark[1] \\
\textbf{Yuki Imajuku} \quad
\textbf{Hikaru Ikuta} \quad
\textbf{Kiyoharu Aizawa} \\
The University of Tokyo \\
\texttt{baek@hal.t.u-tokyo.ac.jp} \\
\url{https://manga109.github.io/MangaVQA_LMM/}
}

\begin{document}
\maketitle

\begin{abstract}
Manga, or Japanese comics, is a richly multimodal narrative form that blends images and text in complex ways. Teaching large multimodal models (LMMs) to understand such narratives at a human-like level could help manga creators reflect on and refine their stories. To this end, we introduce two benchmarks for multimodal manga understanding: MangaOCR, which targets in-page text recognition, and MangaVQA, a novel benchmark designed to evaluate contextual understanding through visual question answering. MangaVQA consists of 526 high-quality, manually constructed question-answer pairs, enabling reliable evaluation across diverse narrative and visual scenarios. Building on these benchmarks, we develop MangaLMM, a manga-specialized model finetuned from the open-source LMM Qwen2.5-VL to jointly handle both tasks. Through extensive experiments, including comparisons with proprietary models such as GPT-4o and Gemini 2.5, we assess how well LMMs understand manga. Our benchmark and model provide a comprehensive foundation for evaluating and advancing LMMs in the richly narrative domain of manga.
\end{abstract}

\section{Introduction}
\label{sec:intro}
Manga is a rich and distinctive form of multimodal narrative, combining complex panel layouts, expressive visual elements, and text embedded directly within images. As large multimodal models (LMMs) continue to advance in vision-language understanding, enabling them to understand manga presents an exciting opportunity, not only as a technical milestone, but also as a way to support human creativity.
Such models could assist manga creators in reflecting on and refining their stories. To provide meaningful assistance, an LMM would need to function like a skilled editor or assistant, capable of reading and understanding manga in a way human does. This calls for evaluating models' abilities to process visual-textual content and follow the context in a coherent and human-like manner.

Although recent efforts such as Magi~\cite{magiv1,magiv2,magiv3} and CoMix~\cite{vivoli2024comix} have tackled comic understanding, they primarily focus on generating transcriptions from comic pages -- they do not evaluate to what extent models can accurately read in-page text using optical character recognition (OCR), or understand the content based on that text through visual question answering (VQA). As a result, it remains unclear to what extent models truly comprehend manga content in a human-like manner based on the embedded textual information.

\begin{figure*}[t]
  \centering
  \includegraphics[width=\linewidth]
  {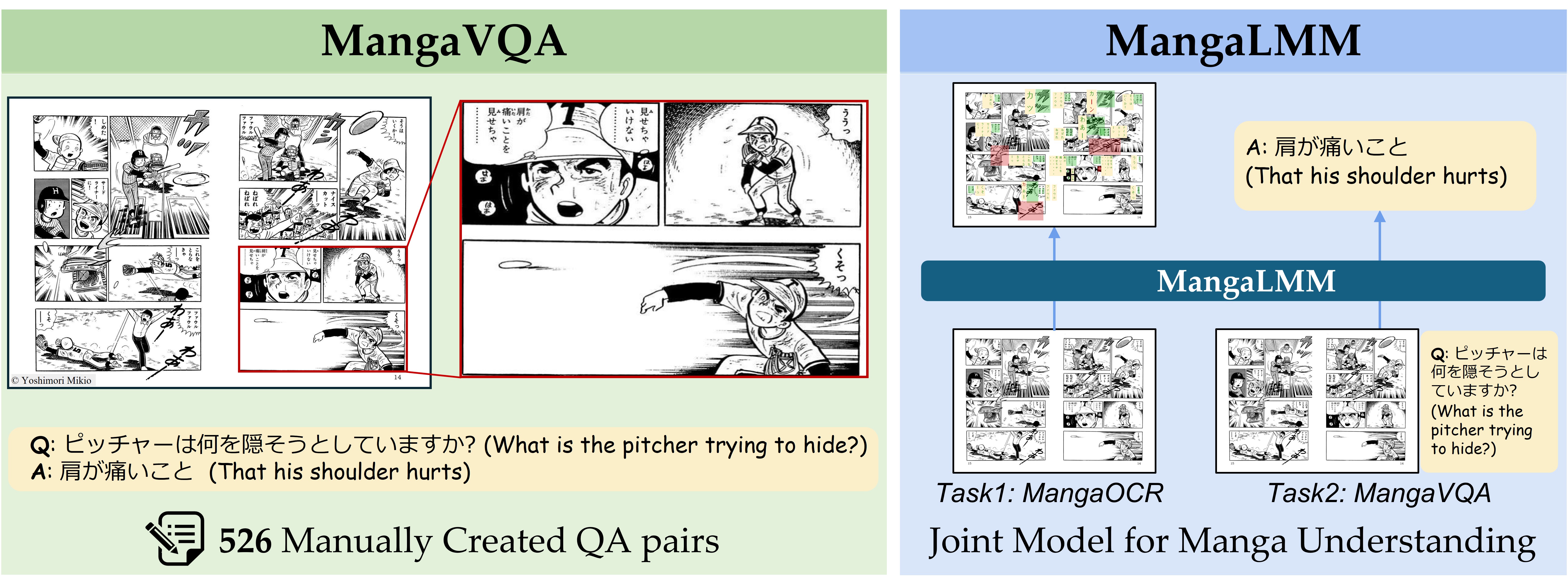}
  \vspace{-8mm}
  \caption{
    \textbf{Overview of MangaVQA and MangaLMM.}  
    We present MangaVQA, a newly proposed benchmark for multimodal context understanding, consisting of 526 manually constructed question–answer pairs. 
    We also develop MangaLMM, a manga-specialized model jointly trained to handle both MangaOCR and MangaVQA tasks.
  }
  \label{fig:overview}
\end{figure*}

To pave a reliable path toward comprehensive manga understanding in LMMs, we believe it is essential to evaluate two core capabilities: OCR and VQA.
To address these needs, we propose two benchmarks: MangaOCR and MangaVQA.
\textbf{MangaOCR} focuses on detecting and recognizing textual content such as dialogue and sound effects. We consolidate existing annotations from the well-known Manga109 dataset~\cite{manga109,manga109_2} and the manga onomatopoeia dataset~\cite{baek2022COO} to construct this benchmark.
Further, as our primary contribution, we propose \textbf{MangaVQA}, a novel benchmark designed to evaluate an LMM's ability to accurately answer targeted, factual questions grounded in both visual and textual context. It consists of 526 high-quality, manually constructed question–answer pairs covering a diverse range of scenarios, enabling assessment of a model's narrative understanding.
Together, these benchmarks provide a comprehensive framework for evaluating a model's ability to understand manga as a multimodal narrative medium, with MangaVQA playing a central role in assessing deeper semantic and contextual comprehension.

Furthermore, truly human-like understanding of manga requires the ability to jointly perform both OCR and VQA, rather than treating them as isolated tasks.
Therefore, building on our two proposed benchmarks, we finetune an open-source LMM (Qwen2.5-VL~\cite{Qwen2.5-VL}) to develop \textbf{MangaLMM}, a manga-specialized model designed to jointly address both OCR and VQA tasks.
MangaLMM serves as a practical baseline for human-like manga understanding.
We conduct comprehensive experiments, including analyses on model and dataset size, and compare MangaLMM with state-of-the-art proprietary models such as GPT-4o~\cite{gpt4o} and Gemini 2.5~\cite{gemini2.5} to evaluate the current landscape of multimodal manga understanding. Our results show that even the proprietary models struggle on our two benchmarks, while MangaLMM jointly handle OCR and VQA, achieving promising performance on both.

An overview of our MangaVQA benchmark and the MangaLMM model is shown in Figure~\ref{fig:overview}.
Our contributions are summarized as follows:
\begin{itemize}[noitemsep, topsep=0pt]
  \item We present \textbf{MangaVQA}, a benchmark of 526 manually constructed question–answer pairs, and \textbf{MangaOCR}, focusing on in-page text detection and recognition. Together, they enable comprehensive evaluation of multimodal manga understanding.

  \item We develop \textbf{MangaLMM}, a manga-specialized version of Qwen2.5-VL, finetuned to jointly address VQA and OCR.

  \item We perform extensive analysis and evaluate MangaLMM against proprietary models, highlighting the limitations of general-purpose LMMs in stylized visual domains.
\end{itemize}

\section{Related Work: Comic Data and Tasks}\label{sec:related}
Recent work, CoMix~\cite{vivoli2024comix}, has unified various comic-related tasks by analyzing existing datasets, including French comics (eBDtheque~\cite{eBDtheque2013}), American comics (COMICS~\cite{Comic-2017amazing} and DCM772~\cite{nguyen2018digital}), and Japanese comics (Manga109~\cite{manga109} and PopManga~\cite{magiv1}). CoMix primarily focuses on transcript generation-related tasks, including object detection, speaker identification, character re-identification, reading order prediction, and character naming prediction.
Similarly, the recent Magi series (v1~\cite{magiv1}, v2~\cite{magiv2}, and v3~\cite{magiv3}) also centers on transcript generation. Notably, Magi v3 extends this pipeline by generating image captions from transcriptions and further producing prose based on those captions.

Although recent studies such as CoMix and the Magi series have addressed a wide range of tasks, the evaluation of OCR has often been underexplored, particularly in detecting the locations of texts within an image and recognizing their content.
One exception is COMICS TEXT+~\cite{soykan2024comprehensive}, which evaluates OCR performance at the panel level, but it does not address page-level evaluation. 
However, humans typically perceive and interpret text at the page level, integrating visual and textual cues across the entire layout.
To reflect this human reading process, we evaluate OCR performance on two-page spreads using MangaOCR. 
A detailed discussion is provided in Appendix~\ref{sup-sec:OCR-eval}.

Existing studies have also largely overlooked the visual question answering (VQA) task in the context of comics. Among prior datasets, the Manga Understanding Benchmark (MangaUB~\cite{mangaub2025}) is the most closely related to our proposed MangaVQA. While MangaUB can be considered a simple VQA benchmark, it contains only eight predefined question types—such as identifying the number of characters, the weather, or the time of day—thus offering limited question diversity. As a result, MangaUB does not address a broad spectrum of VQA problems centered on text understanding in manga. Furthermore, its scope is restricted to the panel level.

In contrast, MangaVQA goes beyond individual panels and focuses on two-page spreads, reflecting how humans naturally read manga. It features diverse VQA questions grounded in textual content at the spread level, aiming to approximate the reading experience of human readers. In this regard, MangaVQA is conceptually aligned with TextVQA~\cite{TextVQA} and DocVQA~\cite{DocVQA}, as it requires models to understand and reason over text embedded in images.

\section{The Manga109 Dataset and Our Consolidated MangaOCR Dataset}\label{sec:manga109andOCR}
This section presents the widely used manga dataset Manga109~\cite{manga109} and our MangaOCR Benchmark.

\subsection{Manga109: A Widely Used Dataset for Manga Research}
Among the many comic datasets introduced in the Related Work, we selected Manga109 for its open-access license, diverse manga titles, and rich annotations and meta-information. It has also been widely used in previous comic-related research~\cite{magiv2,magiv3,baek2022COO,li2024zeroshot,mangaub2025}, making it a reliable and practical dataset for our study.

\begin{figure}[t]
  \centering
  \includegraphics[width=\linewidth]{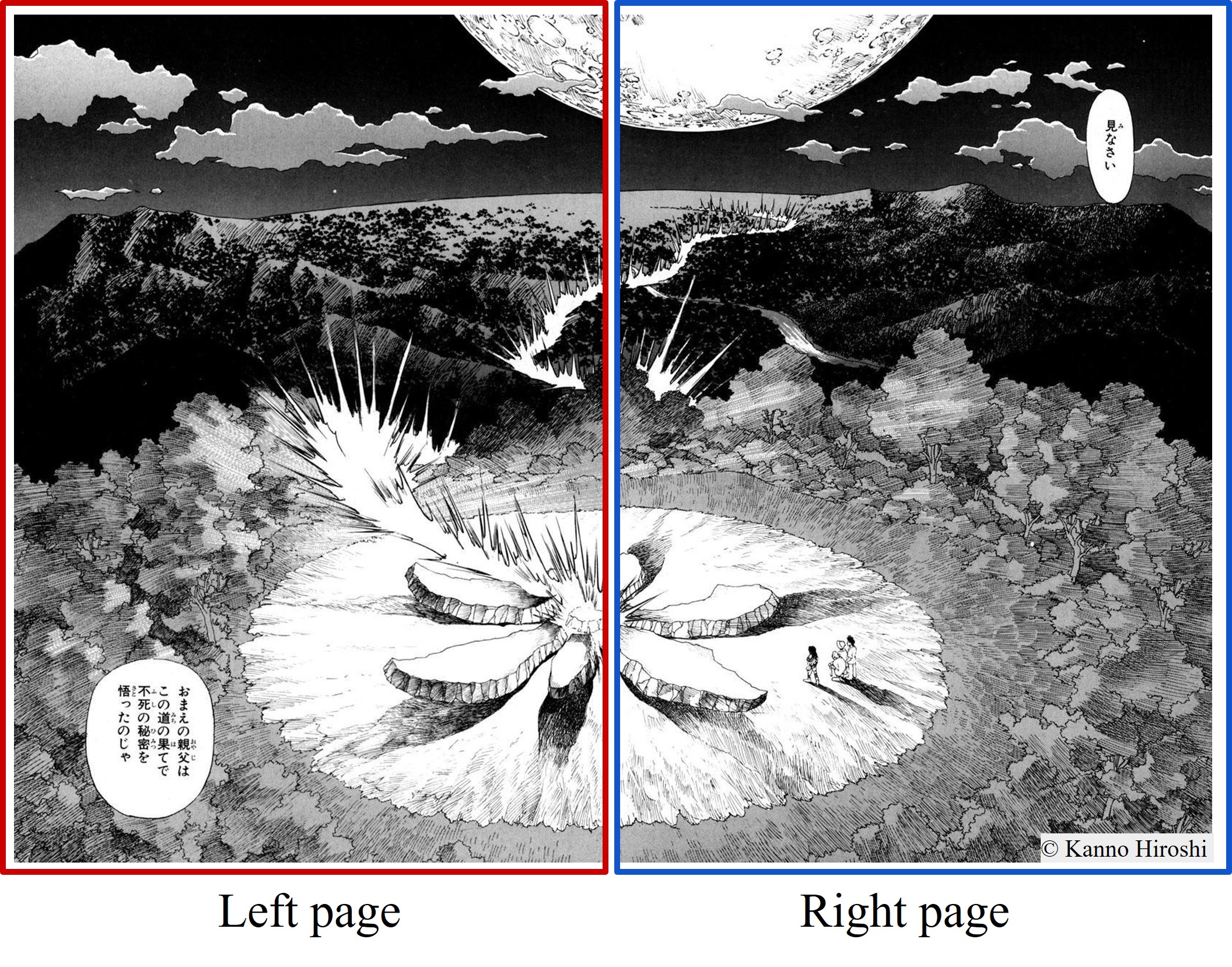}
  \vspace{-8mm}
  \caption{
    Illustration of a two-page spread from the Manga109 dataset.
  }
  \label{fig:Manga109}
  \vspace{-2mm}
\end{figure}

Manga109 is a dataset composed of 109 volumes of Japanese comics (manga). 
Manga is a unique visual storytelling medium characterized by spatially arranged panels and artistic expression. 
The Manga109 dataset captures many distinctive features of manga, including its predominantly black-and-white artwork, two-page spreads, right-to-left reading order, vertical text layout, and the frequent use of stylized onomatopoeia (e.g., Boom, Bang) integrated into the illustrations. 
It also contains culturally specific dialogue, often incorporating honorifics and idiomatic expressions. 
Although these characteristics are not explicitly annotated, they present unique challenges for manga understanding tasks. 
Given these characteristics, Manga109 serves as a representative dataset for developing and evaluating manga understanding models.
Figure~\ref{fig:Manga109} shows an example of two-page spreads from the Manga109 dataset.

\begin{figure*}[t]
\centering
    \subfloat[Required Information]{
    \includegraphics[width=0.23\linewidth]{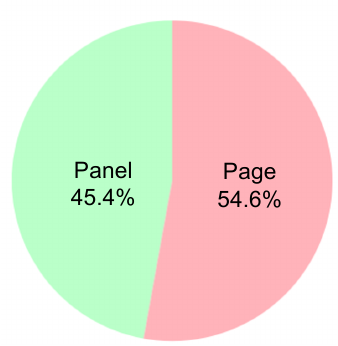}%
  }\hfill
  \subfloat[Answer Type]{
    \includegraphics[width=0.23\linewidth]{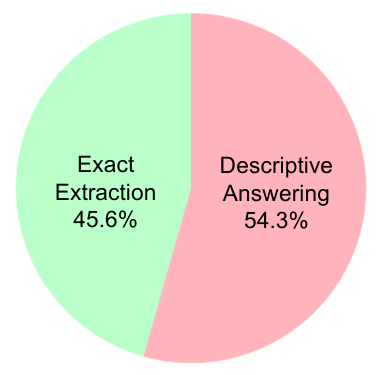}%
  }\hfill
  \subfloat[5W1H]{
    \includegraphics[width=0.23\linewidth]{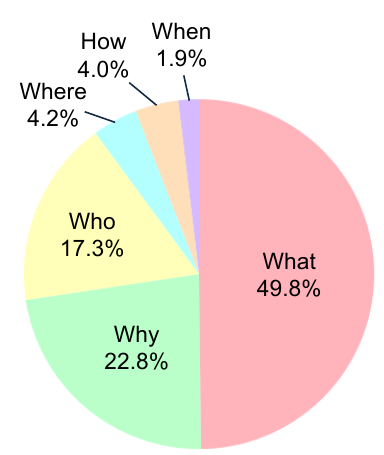}%
  }\hfill
  \subfloat[Author Type]{
    \includegraphics[width=0.23\linewidth]{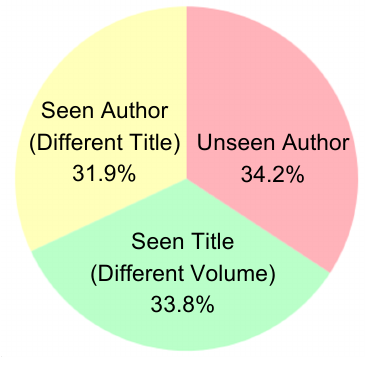}
  }
  \vspace{-2mm}
  \caption{\textbf{Distributions in MangaVQA.} The dataset is structured along four key axes: (a) Required Information, (b) Answer Type, (c) 5W1H, and (d) Author Type.}
    \label{fig:Distribution}
 \vspace{-2mm}
\end{figure*}

\subsection{MangaOCR: A Consolidated Dataset for Manga Text Recognition}\label{subsec:mangaocr}
Text in manga carries essential narrative information, appearing as speech balloons and stylized onomatopoeia integrated into the artwork.  
Recognizing such text is crucial for machine understanding of manga, as humans also rely on this information to comprehend the story.  
MangaOCR addresses this challenge by targeting two key categories of embedded text: dialogue and onomatopoeia.  
We construct the MangaOCR dataset by consolidating existing annotations from the Manga109 dataset and the manga onomatopoeia dataset~\cite{baek2022COO}.  
It contains approximately 209K narrative text instances, spanning a wide variety of visual styles and layouts.  
Training with MangaOCR can improve the ability of LMMs to extract and interpret textual information in manga, contributing to better overall understanding.
The MangaOCR task is performed on two-page spreads and primarily consists of two sub-tasks: text detection, which localizes textual regions, and text recognition, which reads the localized text.

\begin{table}[t]
    \tabcolsep=0.13cm
    \centering
     \begin{tabular}[t]{@{}lrrrr@{}}
        \toprule
        \textbf{Count type} & \textbf{Total} & \textbf{Train} & \textbf{Valid} & \textbf{Test} \\
        \midrule
        Comic volumes & 109 & 89 & 7 & 13 \\
        Images & 10,602 & 8,763 & 673 & 1,166 \\
        \midrule
        MangaOCR \\ 
        ~~Dialogue & 148K & 120K & 9K & 18K \\
        ~~Onomatopoeia & 61K & 50K & 4K & 7K \\
        ~~Total & 209K & 170K & 13K & 26K \\
        \midrule
        MangaVQA \\ 
        ~~ QA pairs & 40,363 & 39,837 & $-$ & 526 \\
        \bottomrule
    \end{tabular}
    \vspace{-2mm}
    \caption{
    \textbf{Statistics of manga datasets.}
    More details about MangaVQA are presented in \S\ref{sec:mangavqa} and \S\ref{sec:mangalmm}.
    }
    \label{tab:stat}
    \vspace{-2mm}
\end{table}

\PAR{Author-Aware Dataset Split}
We adopt the dataset split protocol from prior work~\cite{baek2022COO}, with a few modifications. 
In the original split, the 109 volumes were divided into training, validation, and test sets based on author information. 
To evaluate intra-series generalization, five of the ten test volumes belong to the same series as those in the training set, where the first volume is included in the training set and the last volume is in the test set. 
This setting tests whether a model trained on the beginning of a series can generalize to its later volumes. 
To evaluate intra-author generalization, the remaining five test volumes are titles by authors who also have other works in the training set. 
This allows us to assess whether a model can generalize across different works by the same author.

To further evaluate out-of-distribution generalization with respect to author identity, we move three volumes from the validation set to the test set. These volumes are authored by individuals who did not contribute to any works in the training set.  
Table~\ref{tab:stat} shows the dataset statistics after the split.

\begin{figure}[t]
   \centering
   \includegraphics[width=0.9\linewidth]{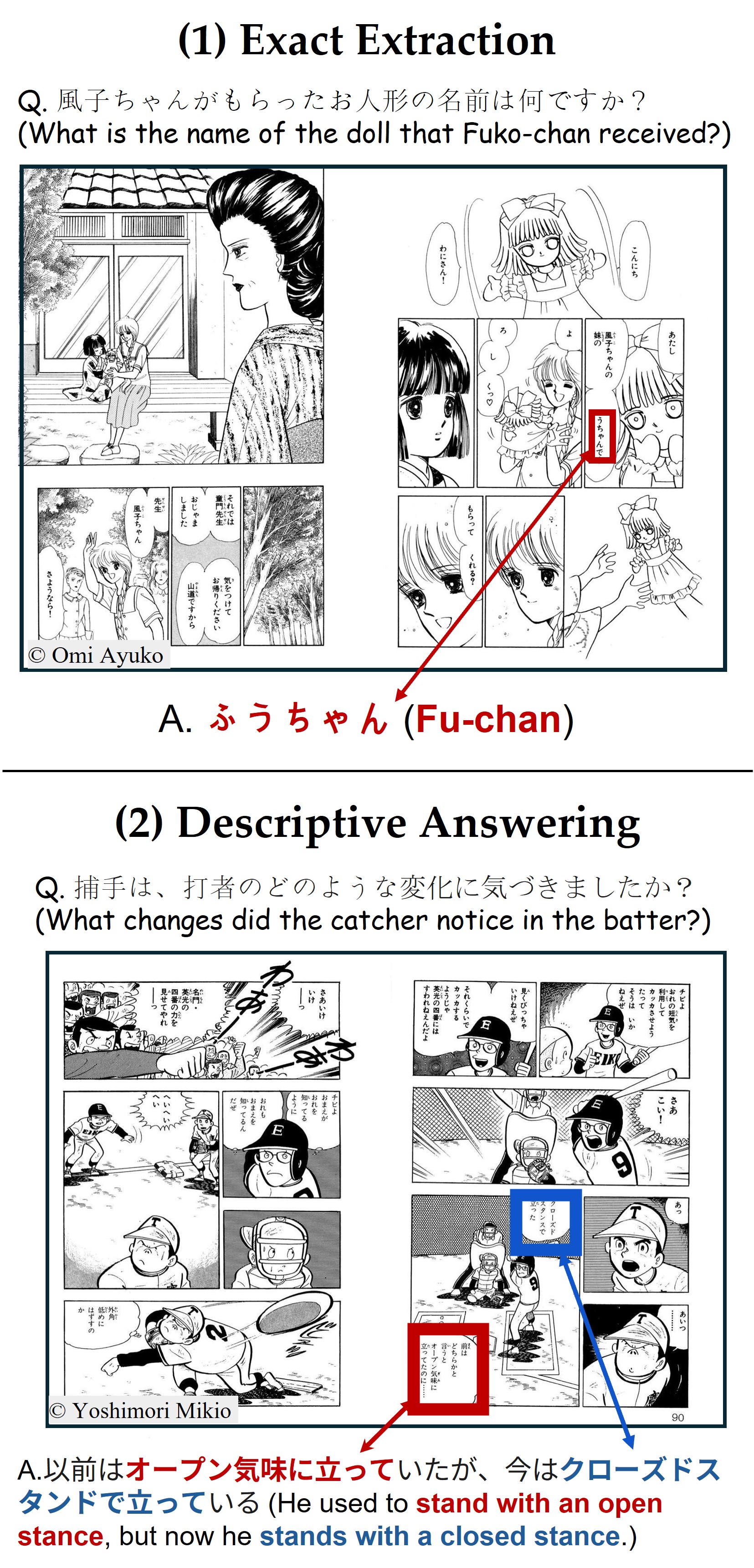}
   \vspace{-4mm}
   \caption{
   \textbf{Main categorization of MangaVQA: Answer type.}
   MangaVQA consists of (1) Exact Extraction, where the answer is directly extracted from the image; and (2) Descriptive Answering, where the answer requires explanatory or contextual responses beyond simple word extraction.
   }
   \label{fig:MangaVQA}
   \vspace{-2mm}
\end{figure}

\section{MangaVQA: A Novel Benchmark for Multimodal Context Understanding}\label{sec:mangavqa}
To evaluate model performance under realistic conditions, we manually constructed a set of question–answer (QA) pairs based on images from Manga109. 
Five annotators among the authors carefully and meticulously developed a high-quality evaluation set for MangaVQA, incorporating thorough human inspection and verification. 
To ensure robust and unambiguous evaluation, we focused on questions with definite answers, avoiding those that could be inferred merely from vague visual impressions. 
The 526 well-curated samples offer a comprehensive evaluation signal that sufficiently covers various aspects for reliable model comparison.

As shown in~\Cref{fig:Distribution}, the question types are designed based on four key axes:
(a) whether solving the question requires information from a single panel or multiple panels at the page level,
(b) the answer type, distinguishing between exact extraction (word-level answers) and descriptive answering (sentence-level or explanatory answers),
(c) 5W1H: whether the question asks about a person (who), an object or action (what), a time (when), a place (where), a reason (why), or a method or condition (how), and
(d) inclusion of the author / title in the training split.

We illustrate examples along axis (b), the answer type, in Fig.~\ref{fig:MangaVQA}.
The categorization of (b) the answer type is as follows:
\textbf{(1) Exact Extraction (240 questions): Questions that Require Extracting Answer Words from the Image.} 
These questions necessitate accurately retrieving the answer word from the manga page. We include one example in the left of Fig.~\ref{fig:MangaVQA}.
The question is ``What is the name of the doll that Fuko-chan received?'' and the answer is ``Fu-chan'', which is directly written in the dialogue. 
This category assesses the LMM's basic comprehension ability to identify and extract the correct answer part from the manga panels. 

\textbf{(2) Descriptive Answering (286 questions): Questions that Require Contextual or Explanatory Responses.}
These questions go beyond simple answer word extraction and require comprehending the context within the manga. 
We include one example in the middle of  Fig.~\ref{fig:MangaVQA}. 
The question is
``What changes did the catcher notice in the batter?''. 
The correct answer is
``He used to stand with an open stance, but now he stands with a closed stance''.
This category allows us to evaluate whether the LMM can not only recognize the dialogue but also understand its underlying meaning in the context of the narrative.

\section{MangaLMM: A Specialized Model for MangaOCR and MangaVQA}\label{sec:mangalmm}
We develop MangaLMM, a specialized model designed to read and understand manga in a human-like manner.
To build MangaLMM, we finetune the open-source LMM Qwen2.5-VL~\cite{Qwen2.5-VL} on the MangaOCR and MangaVQA datasets, resulting in a joint model for both tasks.
In this section, we describe the training data construction and training details for MangaLMM.

\subsection{Training Data Construction}
\PAR{OCR Training Set $\mathrm{T_{OCR}}$}
For the OCR task, we use the MangaOCR training set, as described in \S\ref{subsec:mangaocr}. 
For each image, we format the sequence of text annotations as 
\texttt{\{"bbox\_2d":coordinates$_i$, "text\_content":text$_i$\}}, for each $i$, where \texttt{coordinates$_i$} corresponds to the location of the \texttt{text$_i$} in the image represented as \texttt{x$_\text{top\_left}$,y$_\text{top\_left}$,x$_\text{bottom\_right}$,y$_\text{bottom\_right}$}.

\PAR{Synthetic VQA Training Set $\mathrm{T_{VQA}}$}
For the VQA task, we generate synthetic training data using GPT-4o~\cite{gpt4o} (\texttt{gpt-4o-2024-11-20}).
Following the synthetic data construction used in LLaVA~\cite{llava}, we generate five questions per image using both the image and its annotation from the OCR training set $\mathrm{T_{OCR}}$.
Here we exclude < 0.1\% of the images where the text annotation is not included or GPT-4o refused to respond (e.g., due to violent content).
Although we requested GPT-4o to generate five questions per image, it occasionally returned fewer than five.
As a result, we created a total of 39,837 synthetic VQA samples from 8,379 images. 
The prompt used for question generation is shown in Table~\ref{sup:prompt-genVQA} in the Appendix.
We plan to release this as a training split of our MangaVQA.

\subsection{Training Details}
\PAR{LMM Selection}
Our tasks require an open-source multilingual LMM that can handle Japanese and also has strong Japanese OCR capabilities, which are important for understanding manga. Several powerful multilingual LMMs have been proposed recently~\cite{yue2024pangea, wang2024qwen2, Qwen2.5-VL, maaz2024palo, coherelabs2025, phi4-2025}. Among them, the Qwen series~\citep{wang2024qwen2, Qwen2.5-VL} and Phi-4~\citep{phi4-2025} are especially notable for their Japanese OCR performance. In this work, we build MangaLMM based on Qwen2.5-VL~\citep{Qwen2.5-VL}, which is one of the strongest open-source models in this category.

\PAR{Training Strategy}
We perform continual finetuning on both $\mathrm{T_{OCR}}$ and $\mathrm{T_{VQA}}$ using the pretrained Qwen2.5-VL 7B (\texttt{Qwen2.5-VL-7B-Instruct}).  
Most hyperparameters follow the original Qwen2.5-VL configuration, with a few modifications.  
For Manga109 images (1654$\times$1170 resolution), we follow Qwen2.5-VL's image resizing mechanism, which is based on pixel count thresholds, where the minimum and maximum number of input pixels are 3,136 and 2,116,800, respectively.
We train for one epoch with a batch size of 32.

\PAR{Elapsed Time for Training} 
Each dataset is trained for one epoch. 
Training Qwen2.5-VL 7B using four NVIDIA A100 GPUs took about 1 hour when using $\mathrm{T_{OCR}}$ or $\mathrm{T_{VQA}}$, and about 2 hours when using both $\mathrm{T_{OCR}}$ and $\mathrm{T_{VQA}}$.

\section{Experiments}\label{sec:experiment}
\PAR{Evaluation Protocol for MangaOCR}
We follow the evaluation protocols from prior OCR studies~\cite{DeepSolo,ESTextSpotter} and ICDAR 2019 multilingual OCR competitions~\cite{ArT,ReCTS,LSVT,MLT19}.
First, a predicted bounding box is considered a correct detection if its intersection over union (IoU) with a ground truth box exceeds 0.5. Based on the matched boxes, we compute precision (P), recall (R), and the harmonic mean (Hmean).
Second, for each matched box, we calculate the normalized edit distance (NED) between the predicted and ground truth texts as a character-level metric.
NED ranges from 0 to 1, with higher values indicating better performance; details are in the supplementary materials.

Since LMMs sometimes output the same word repeatedly, we apply post-processing to exclude repeated text segments that appear more than ten times, treating them as noise.
Except for the analysis in \S\ref{subsec:Analysis-MangaOCR}, we report only the end-to-end Hmean for simplicity.

\PAR{Evaluation Protocol for MangaVQA}
Following LLaVA-Bench~\cite{llava}, we adopt the LLM-as-a-judge approach~\cite{judge} as our evaluation metric. 
We select a single LLM as the judge, which assigns scores to model responses. 
Specifically, we provide the judge with the question, a human-written answer, and the model's response. 
Based on the human-written answer, the judge evaluates whether the model's response is appropriate and relevant to the question on a 1–10 scale. 
To avoid circular bias with GPT-4o, which was used in constructing our VQA training dataset, we employ Gemini 2.5 Flash~\cite{gemini2.5} (\texttt{gemini-2.5-flash}) as the judge LLM.
The prompt used for LLM-as-a-judge is shown in Table~\ref{sup:prompt-VQAeval} in the Appendix.

\PAR{LMMs Used for Comparison}
We evaluate three proprietary LMMs, gpt-4o-2024-11-20~\cite{gpt4o}, gemini-2.5-flash~\cite{gemini2.5}, and claude-sonnet-4-5-20250929~\cite{anthropic2025sonnet4_5},
and seven open-source LMMs with Japanese capability: 
Qwen2.5-VL-7B-Instruct~\cite{Qwen2.5-VL}, Phi-4-multimodal-instruct~\cite{phi-4MM}, Pangea-7B~\cite{yue2025pangea}, LLaVA-OneVision-1.5-8B-Instruct~\cite{an2025llava-OV1.5}, 
sarashina2-vision-8b~\cite{sarashina2-vision-8b}, gemma-3-12b-it~\cite{team2025gemma}, 
and Heron-NVILA-Lite-15B~\cite{heron-nvila-lite-15b}.

\subsection{Main Results}\label{subsec:main_results}

\begin{table}[t]
    \tabcolsep=0.13cm
    \centering
    \begin{adjustbox}{width=\linewidth}
     \begin{tabular}[t]{@{}l|c|c@{}}
        \toprule
        & \textbf{MangaOCR} & \textbf{MangaVQA} \\ 
        \textbf{Method} & Hmean (\%) & LLM (/10.0) \\
        \midrule
        GPT-4o & 0.0 & 6.00 \\
        Gemini 2.5 Flash & 0.0 & \textbf{7.26} \\
        Claude Sonnet 4.5 & 0.0 & 5.84 \\
        \midrule
        Phi-4-Multimodal-5.6B & 0.0 & 3.39 \\
        Pangea-7B & 0.0 & 3.23 \\
        LLaVA-OV-1.5-8B & 0.0 & 3.46 \\
        Sarashina2-Vision-8B & 0.0 & 4.45 \\
        Gemma-3-12B & 0.0 & 4.13 \\
        Heron-NVILA-Lite-15B & 0.0 & 3.76 \\ 
        Qwen2.5-VL-7B & 0.9 & 5.65 \\ 
        \midrule
        MangaLMM (Ours) & \textbf{71.5} & 6.68 \\
        \bottomrule
    \end{tabular}
    \end{adjustbox}
    \vspace{-2mm}
    \caption{
    \textbf{Comparison of LMMs on MangaOCR and MangaVQA.}
    }
    \label{tab:main-results}
\end{table}

Table~\ref{tab:main-results} compares LMMs for both MangaOCR and MangaVQA tasks.
Overall, MangaLMM can handle both tasks effectively: it achieves over 70\% OCR score and shows competitive VQA performance. 
While it falls short of the proprietary model Gemini, it outperforms the other proprietary models GPT-4o and Claude Sonnet 4.5.
MangaLMM achieves the best performance among the open-source models by a clear margin.

\PAR{Analysis of Low Performance on MangaOCR} 
As shown in~\Cref{tab:main-results}, all LMMs except MangaLMM show near-zero scores on the MangaOCR benchmark.
Most of their predictions consist of meaningless repetitions or short repeated tokens.
The extremely low OCR score before finetuning is likely due to two main factors:  
(1) these models are not familiar with manga data, and 
(2) their weak detection capabilities may limit OCR performance. 
Prior work~\cite{wu2024dettoolchain} has shown that GPT-4o, for example, exhibits poor detection ability, which may also apply to the other models.

Despite the near-zero OCR score—where not only position information is missing but even the correct text content is not generated—these models still manage to answer certain VQA questions that require interpreting text within the image.
This is somewhat \textit{counterintuitive}.
Although the models fail to explicitly output the correct OCR results, they appear to capture some textual semantics from the image.
This suggests that they are able to extract relevant information needed for answering VQA questions, even without performing OCR correctly.
We provide more analysis and discussion in Appendices~\ref{sup-sec:analysis-mangaocr} and~\ref{sup-sec:OCR-eval}.

\begin{table}[t]
    \tabcolsep=0.13cm
    \centering
     \begin{tabular}[t]{@{}l|c|c@{}}
        \toprule
        & \textbf{MangaOCR} & \textbf{MangaVQA} \\ 
        \textbf{FT data} & Hmean (\%) & LLM (/10.0) \\
        \midrule
        None & 0.9 & 5.65 \\
        $\mathrm{T_{OCR}}$ & \textbf{74.5 $\pm$ 1.3} & 1.20 $\pm$ 0.32 \\
        $\mathrm{T_{VQA}}$ & 0.0 $\pm$ 0.0 & 6.57 $\pm$ 0.06 \\
        $\mathrm{T_{OCR}}$+$\mathrm{T_{VQA}}$ & 71.2 $\pm$ 0.6 & \textbf{6.68 $\pm$ 0.03} \\
        \bottomrule
    \end{tabular}
    \vspace{-2mm}
    \caption{
    \textbf{Effect of finetuning (FT).}
    FT is performed on the OCR training set $\mathrm{T_{OCR}}$, the VQA training set $\mathrm{T_{VQA}}$, or both.
    }
    \label{tab:FT}
\end{table}

\paragraph{Analysis of the Effect of Finetuning.}
Table~\ref{tab:FT} shows the effect of finetuning.
Each experiment was conducted three times, and the mean and standard deviation are reported. 
Finetuning Qwen2.5-VL on $\mathrm{T_{OCR}}$ and $\mathrm{T_{VQA}}$ enables the model to specialize in each respective task. 
On MangaOCR, the finetuned model achieves a significant improvement to a score of 74.5$\pm$1.3, which we discuss further in \S\ref{subsec:Analysis-MangaOCR}.
On MangaVQA, the model, initially underperforming GPT-4o, surpasses it after finetuning.
These results highlight the effectiveness of our synthetic VQA training set $\mathrm{T_{VQA}}$, which we further analyze in \S\ref{subsec:Analysis-MangaVQA}.

\paragraph{Analysis from the Perspective of Task Interference.}
MangaLMM, a Qwen2.5-VL model finetuned jointly on both $\mathrm{T_{OCR}}$ and $\mathrm{T_{VQA}}$, shows a slight drop in OCR performance compared to using $\mathrm{T_{OCR}}$ alone, but achieves a small gain in VQA score over using $\mathrm{T_{VQA}}$ alone. 
A common issue in multi-task learning is \textit{task interference}~\cite{maninis2019attentive,yu2020gradient,ding2023mitigating,chen2024octavius}, where models jointly trained on multiple tasks (e.g., $A$ and $B$) often perform worse on task $A$ than models trained solely on $A$. 
Under this assumption, one might expect the VQA performance of a jointly trained OCR+VQA model to degrade relative to a VQA-only model. 
Interestingly, we instead observe a slight improvement in VQA performance under joint training, contrary to typical interference expectations. 
This suggests that although task interference may exist, the enhanced OCR capability likely provides helpful textual cues that marginally improve VQA performance.

\subsection{Performance Analysis of MangaOCR}\label{subsec:Analysis-MangaOCR}
\Cref{tab:OCR-det-e2e} presents MangaOCR performance at both the detection and end-to-end stages.  
The Hmean of detection is 78.6\%, while that of end-to-end reaches 71.5\%, implying that once text regions are detected, the model reads them with approximately 91.0\% (=71.5 / 78.6) accuracy.  
Some false positives occur when the model predicts text that actually appears in the manga but is not included in the annotations—for instance, page numbers or editorial marks outside the narrative content such as dialogue or onomatopoeia.  
Consequently, precision is unlikely to reach 100\%.  
In contrast, recall is relatively low (68.5\%), suggesting that about 31.5\% of the ground-truth narrative text remains undetected, leaving room for improvement in capturing all semantically relevant content.

\begin{table}[t]
    \centering
    \begin{tabular}[t]{@{}l|ccc@{}}
        \toprule
        \textbf{Stage} & \textbf{Prec.} & \textbf{Recall} & \textbf{Hmean} \\
        \midrule
        Detection & 82.2 & 75.3 & 78.6 \\
        End-to-end & 74.8 & 68.5 & 71.5 \\
        \bottomrule
    \end{tabular}
    \vspace{-2mm}
    \caption{
    MangaLMM's detection and end-to-end performance on MangaOCR.
    }
    \label{tab:OCR-det-e2e}
\end{table}

\begin{figure*}[t]
  \centering
  \includegraphics[width=\linewidth]{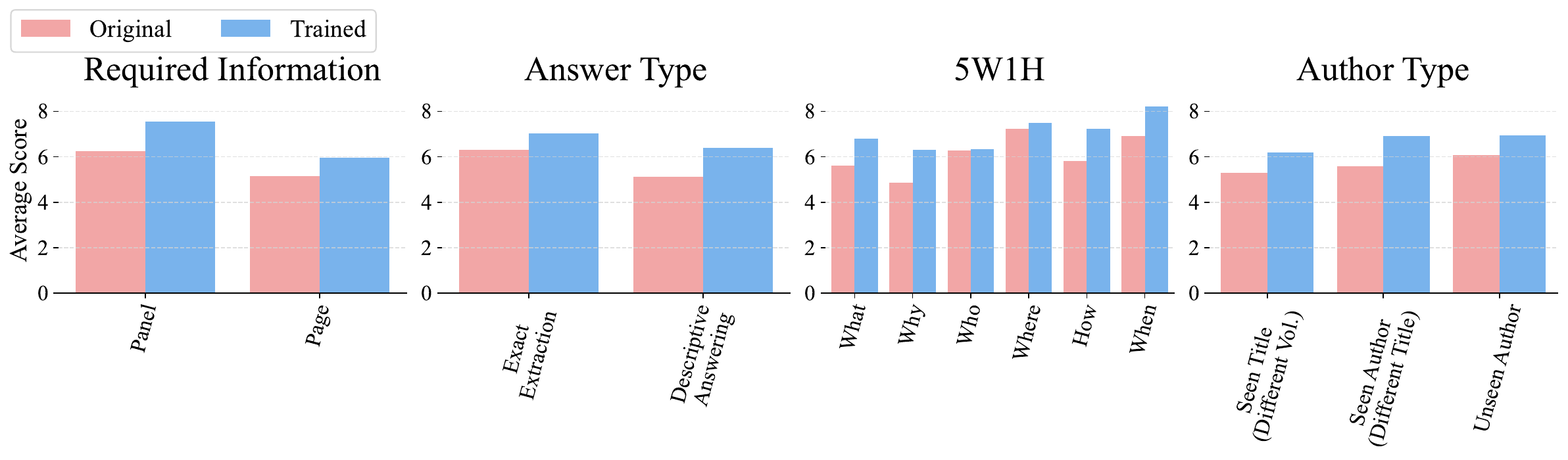}
  \vspace{-8mm}
  \caption{
    \textbf{Category-wise score breakdown.}
    Compared to the original model (Qwen2.5-VL-7B-Instruct), our trained MangaLMM improves scores across every tag in every category.
  }
  \label{fig:score_per_tag}
\end{figure*}

\begin{figure*}[t]
  \centering
  \includegraphics[width=\linewidth]{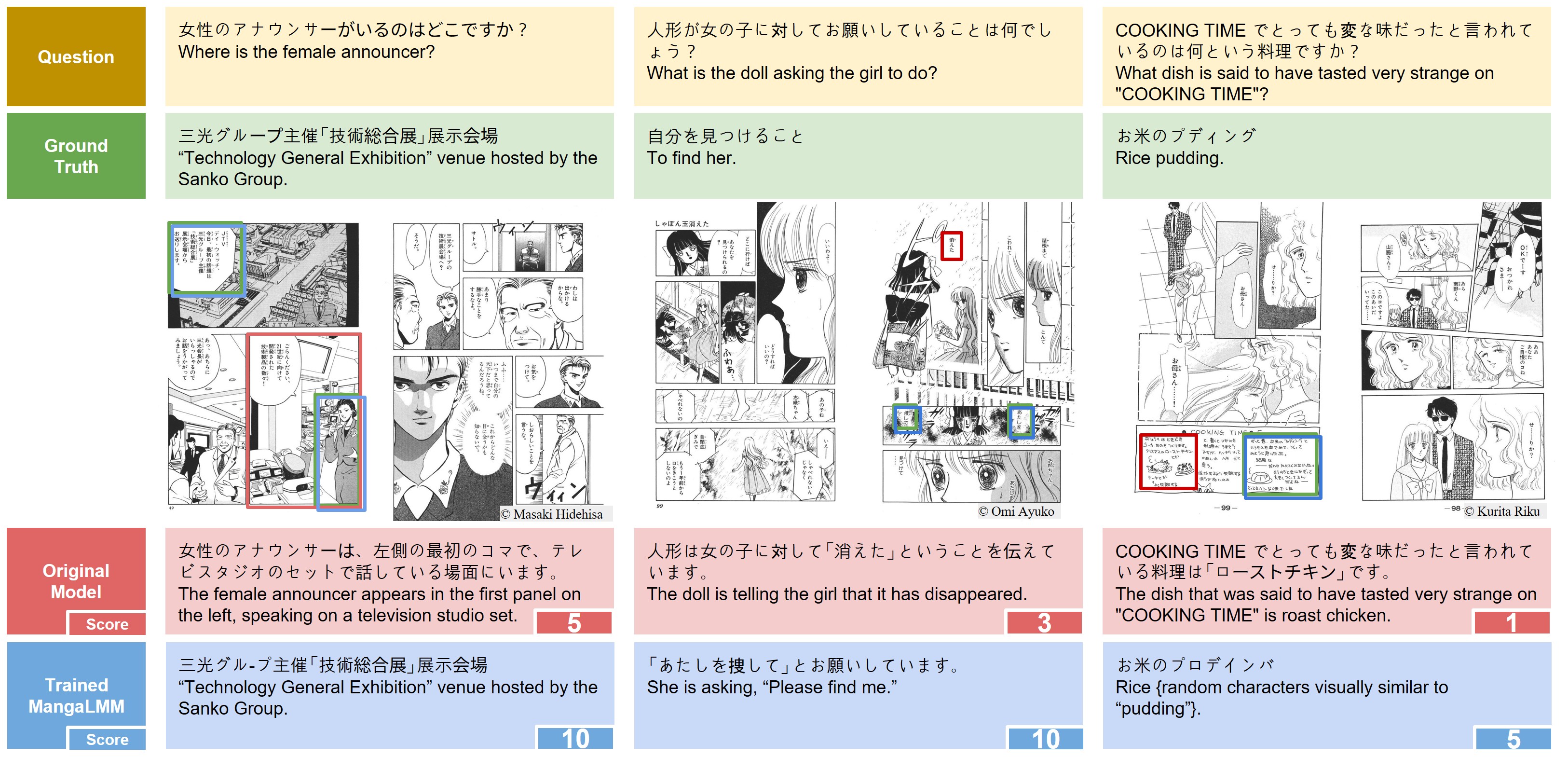}
  \vspace{-8mm}
  \caption{
    \textbf{Qualitative analysis on MangaVQA.}
    Regions relevant to the question or models' answer are highlighted with boxes in corresponding colors.
    In the left and middle examples, performance improves significantly after training, whereas in the right example, the trained model still struggles to produce an accurate answer.
  }
  \label{fig:quali_MangaVQA}
\end{figure*}

\subsection{Performance Analysis of MangaVQA}\label{subsec:Analysis-MangaVQA}
\PAR{Category-wise VQA Performance}
\Cref{fig:score_per_tag} presents a breakdown of model performance across the annotated categories in MangaVQA.
We observe consistent performance gains across all tags, indicating that our training contributes to stable improvement in VQA capability without favoring specific categories.
Interestingly, the model also generalizes well to questions from unseen authors.

\paragraph{Effect of OCR Annotation when Generating VQA Data.}
\begin{table}[t]
    \centering
     \begin{tabular}[t]{@{}c|c@{}}
        \toprule
        \textbf{OCR Annot.} & \textbf{LLM (/10.0)} \\
        \midrule
        & 5.64 \\
        \checkmark & \textbf{6.68} \\
        \bottomrule
    \end{tabular}
    \vspace{-2mm}
    \caption{
    Effect of OCR Annotation on VQA Generation.
    }
    \label{tab:genVQA-OCRGT}
\end{table}

When generating synthetic QA pairs for training, we include the OCR annotations in the prompt provided to GPT-4o.  
To examine their impact, we compare VQA data generated with and without text information.  
As shown in~\Cref{tab:genVQA-OCRGT}, the model trained on VQA data generated without OCR annotations achieves a score of 5.64, which does not exceed GPT-4o's own performance (6.00).  
In contrast, using OCR-guided VQA data significantly improves the score to 6.68, even surpassing GPT-4o.  
These findings suggest that incorporating OCR annotations helps GPT-4o generate higher-quality QA pairs that enable the trained model to surpass GPT-4o's own performance.

\paragraph{Qualitative Analysis of MangaVQA.}
In~\Cref{fig:quali_MangaVQA}, we compare outputs of the original Qwen model and our trained model:
\textbf{Left}: The original model provides a general answer, whereas the trained model leverages text-bubble content to produce a more specific one, improving the score ($5 \rightarrow 10$).
\textbf{Middle}: The original model extracts irrelevant text, while the trained model identifies the correct text, yielding a higher score ($3 \rightarrow 10$).
\textbf{Right}: The original model outputs an unrelated dish name, while the trained model identifies the correct one but makes character-level recognition errors, yielding a partial score increase ($1 \rightarrow 5$).

\section{Conclusion} \label{sec:conclusion}
We presented MangaVQA, a benchmark for evaluating how well LMMs can understand manga in a human-like manner through contextual visual question answering, and MangaOCR, a consolidated benchmark for in-page text recognition.
Together, they cover both the textual and narrative aspects of multimodal manga understanding.
To establish a strong baseline, we developed MangaLMM, a specialized model jointly finetuned on OCR and VQA tasks.
Experiments show that even state-of-the-art proprietary LMMs struggle with the unique complexity of manga, while MangaLMM performs well across both tasks.
By releasing open benchmarks, synthetic data, and a strong open-source baseline, we aim to foster future research on multimodal manga understanding.
Our work provides a concrete example of building and evaluating context-aware, domain-specialized LMMs, serving as a practical reference for similar research in other domains.

\section*{Limitation}\label{sec:limitation}
One limitation of our model lies in its relatively slow inference speed for OCR.
LMMs are inherently slower than dedicated OCR models; for instance, processing 1,166 test images containing 25,651 text instances takes several hours on an A100 GPU.
In contrast, a dedicated OCR model such as DeepSolo~\cite{DeepSolo}, which runs at over 10 FPS, can complete the same task in about two minutes.
This slowdown primarily results from the large number of output tokens and occasional repetition or looping in the generated outputs during inference.

\section*{Acknowledgments}
This work was supported by JSPS KAKENHI Grant Number 24K23882 and by the NVIDIA Academic Grant Program. This research utilized NVIDIA Saturn Cloud.

\bibliography{custom}

\newcommand\beginsupplement{%
        \setcounter{table}{0}
        \renewcommand{\thetable}{\Alph{table}}%
        \setcounter{figure}{0}
        \renewcommand{\thefigure}{\Alph{figure}}%
     }
\beginsupplement
\appendix
In this supplementary material, we provide additional details including (\ref{sup-sec:OCR-eval}) OCR evaluation in comics, (\ref{sup-sec:synthVQA}) synthetic VQA examples, (\ref{sup-sec:MangaVQAexample}) additional MangaVQA examples, (\ref{sup-sec:setup}) setup details, and (\ref{sup-sec:additional-result}) additional results.

\begin{figure*}[t]
  \centering
  \begin{subfigure}{\linewidth}
    \centering
    \includegraphics[width=\linewidth]{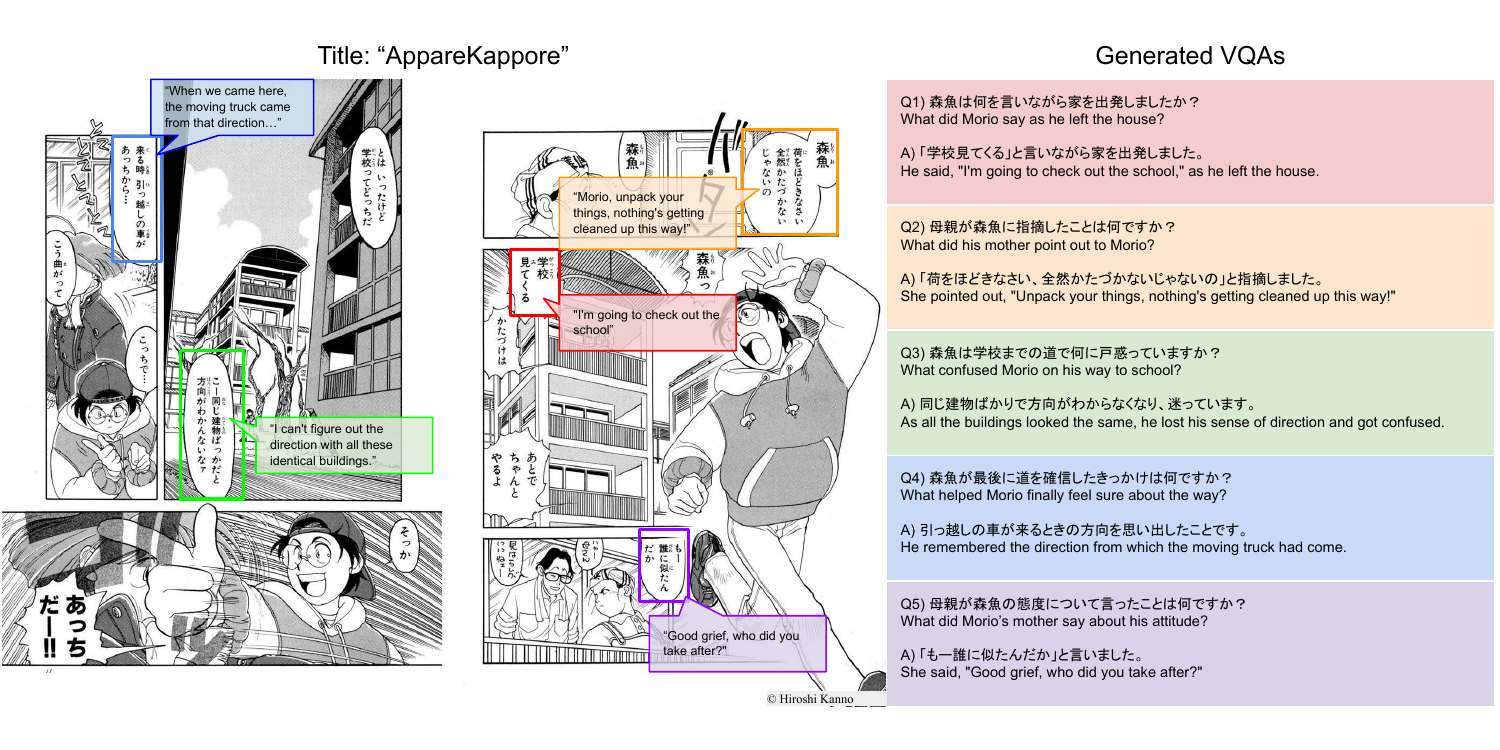}
    \vspace{-6mm}
    \caption{An example from the manga titled AppareKappore.}
    \label{fig:synthetic_vqa_apparekappore}
  \end{subfigure}
  
  \begin{subfigure}{\linewidth}
    \centering
    \includegraphics[width=\linewidth]{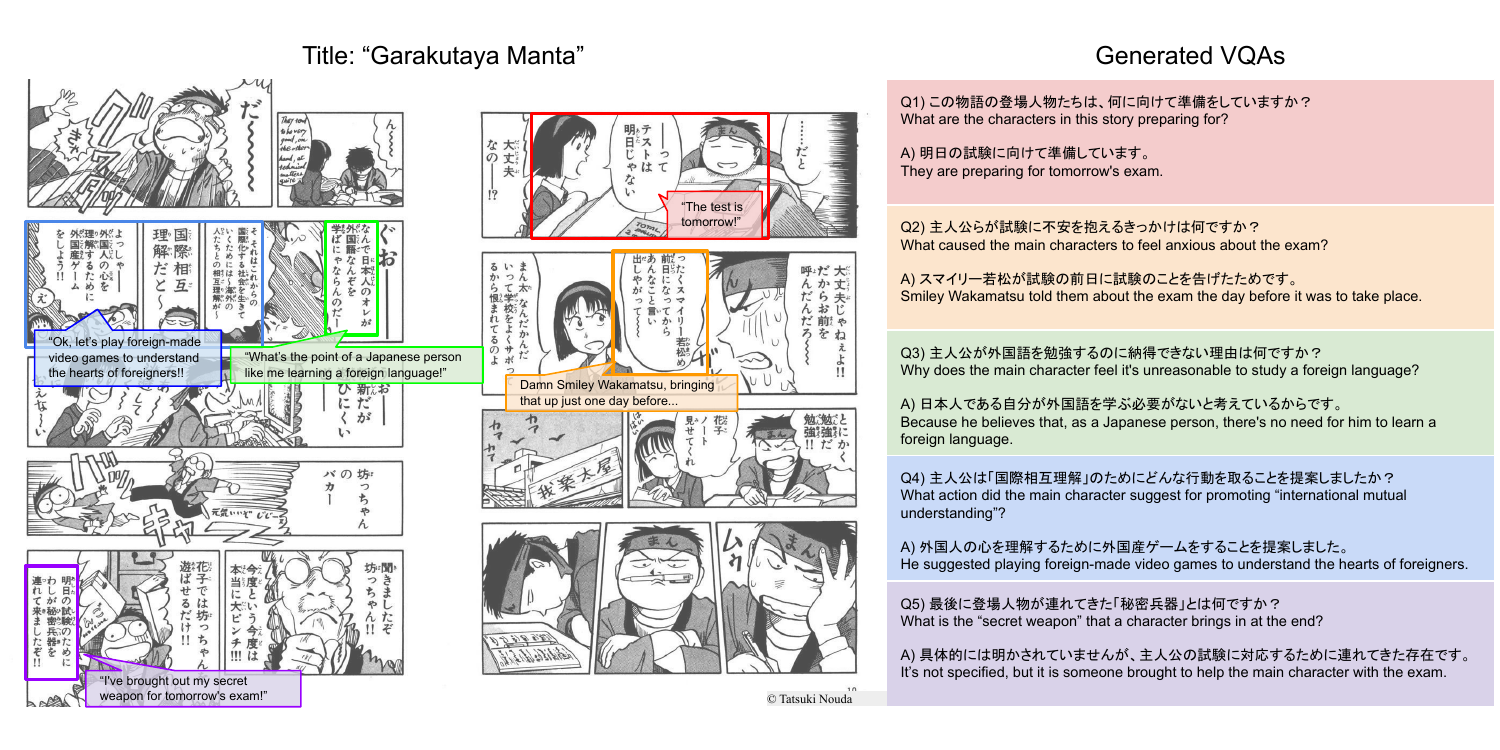}
    \vspace{-6mm}
    \caption{An example from the manga titled GarakutayaManta.}
    \label{fig:synthetic_vqa_garakutayamanta}
  \end{subfigure}
  \caption{
  \textbf{Examples of synthetic VQA generation results.}
  The most relevant part of the image for each question-answer pair is highlighted and translated in the corresponding color.
  }
  \label{fig:synthetic_vqa}
\end{figure*}

\section{OCR Evaluation in Comics}\label{sup-sec:OCR-eval}
As described in \S\ref{sec:related}, the evaluation of OCR has often been underexplored.
Recent works such as Magi~\cite{magiv1} and CoMix~\cite{vivoli2024comix} focus on transcription generation, which inherently includes OCR as a core component. 
CoMix, in particular, proposes a dedicated metric called the Hybrid Dialog Score for evaluating transcription tasks. 
However, this transcription-focused evaluation differs from direct OCR evaluation, which aims to assess whether the model accurately reads the text.
First, transcription involves multiple subtasks beyond text detection and recognition, such as speaker identification, reading order prediction, and others. 
The quality of the final transcription output depends on the combined performance of these components, making it difficult to isolate and measure the accuracy of text recognition alone.

Second, transcription-based evaluations do not assess the positional accuracy of recognized text. 
Spatial information plays a crucial role in OCR, especially when the same text appears in multiple locations, as it helps identify which text instance is correct.
For example, in \Cref{sup-fig:MangaLMM-ocr}(a), the word ``わあー (waa-)'' appears in four different locations, only one of which is correct.
Without positional information, it becomes impossible to identify the correct instance. 
Moreover, spatial information is crucial for content understanding, as the interpretation of the same text can vary significantly depending on its location.

A proper evaluation of OCR in the manga domain allows us to better understand how well current LMMs can recognize text within manga.
As described in the results section (\S\ref{subsec:main_results}), models such as GPT-4o exhibit near-zero OCR performance, yet are still able to answer VQA questions that rely on textual information. 
This result suggests that LMMs may be partially recognizing some text in the image.  
Our visualization of GPT-4o's OCR output reveals that the detected text regions almost always appear in nonsensical locations, yet the model can still read certain parts of the text in the image.
We provide a detailed analysis of this observation in \S\ref{sup-sec:quali-OCR}.

\section{Synthetic VQA Examples}\label{sup-sec:synthVQA}
For training our MangaLMM, we rely on synthetic VQAs generated by GPT-4o.
In~\Cref{fig:synthetic_vqa}, we provide examples of these generated VQAs. As illustrated in the figure, GPT-4o is capable of producing accurate and diverse question–answer pairs.

We emphasize once again that providing GPT with text annotations is crucial for generating such high-quality VQAs. 
Without these annotations, GPT tends to produce unreliable outputs (e.g., misspelled extractions and factually incorrect questions) which significantly limit the performance of the MangaLMM trained on such data, as discussed in \S\ref{subsec:Analysis-MangaVQA}.

\PAR{Human Validation}
To validate the reliability of the synthetic VQA data generated by GPT, we conducted a manual evaluation. We randomly sampled 500 question–answer pairs and asked four human evaluators to assign scores to each item on a three-level scale: 0 (incorrect), 0.5 (partially correct), and 1 (correct). The average score is 0.80, suggesting that approximately 80\% of the synthetic VQAs are judged to be appropriate by humans.

\section{Additional MangaVQA Examples}\label{sup-sec:MangaVQAexample}
\begin{figure*}[t]
   \centering
   \includegraphics[width=0.8\linewidth]{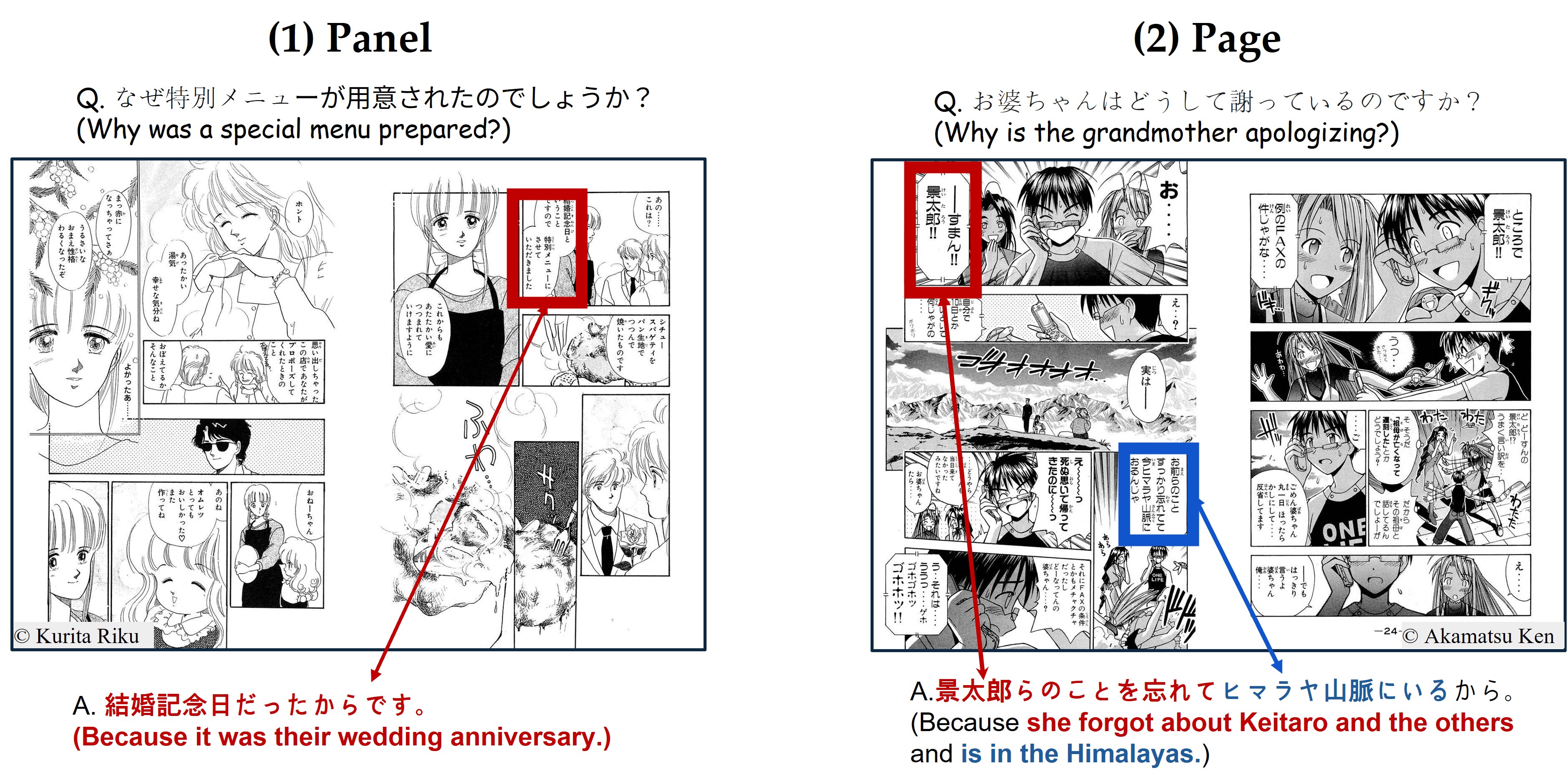}
   \caption{
   \textbf{Categorization of MangaVQA: Required information.}
   MangaVQA questions are categorized by whether solving the question requires information from (1) a single panel or (2) multiple panels at the page level.
   }
   \label{sup-fig:MangaVQA-required_information}
\end{figure*}

\begin{figure*}[t]
   \centering
   \includegraphics[width=\linewidth]{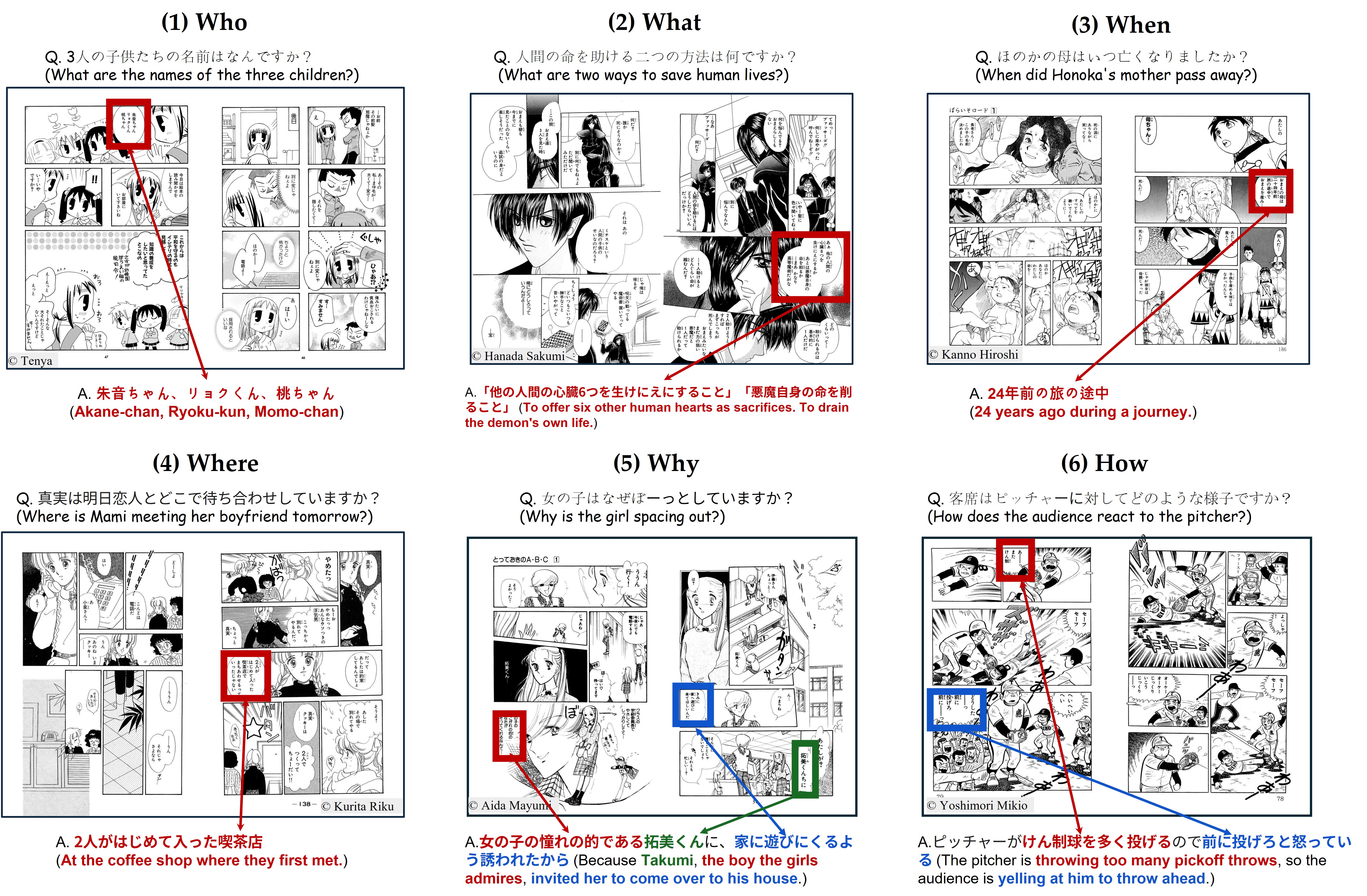}
   \caption{
   \textbf{Categorization of MangaVQA: 5W1H.}
   MangaVQA questions are also categorized by 5W1H, that is, whether the question asks about a person (who), an object or action (what), a time (when), a place (where), a reason (why), or a method or condition (how).
   }
   \label{sup-fig:MangaVQA-5W1H}
\end{figure*}

In this section, we provide additional examples from MangaVQA to illustrate how questions are categorized according to different aspects.
\Cref{sup-fig:MangaVQA-required_information} shows examples categorized by required information, indicating whether the question can be answered from a single panel or requires multiple panels at the page level.
\Cref{sup-fig:MangaVQA-5W1H} presents examples categorized by 5W1H, showing the diversity of question types such as who, what, when, where, why, and how.
These examples highlight the variety of question types and contextual understanding needed to answer MangaVQA questions.

\clearpage

\clearpage

\section{Setup Details}\label{sup-sec:setup}

\PAR{Evaluation Metric}
We provide a detailed description of the normalized edit distance (NED, also referred to as 1-NED), which was used as the evaluation metric in MangaOCR.
NED scales the standard edit distance to a range between 0 and 1, where higher values indicate better prediction.
It is computed as follows:
\begin{align}
\mathrm{NED} &= 1 - \sum^{N}_{i=1}\frac{\mathrm{ED}(\mathrm{GT}_i, \mathrm{Pred}_i)}{\mathrm{MaxLen}(\mathrm{GT}_i, \mathrm{Pred}_i)}
\end{align}
Here, $\mathrm{GT}_i$ and $\mathrm{Pred}_i$ denote the $i$-th ground truth and the model's prediction, respectively.
$\mathrm{ED}(\cdot)$ calculates the edit distance between two strings, and $\mathrm{MaxLen}(\cdot)$ returns the longer of the two string lengths.
$N$ indicates the total number of text instances.

\subsection{Prompt}
\PAR{Prompt for Synthetic VQA Generation}
For creating synthetic QA pairs for training, we provide GPT-4o with the prompt in~\Cref{sup:prompt-genVQA} along with the corresponding image.

\begin{table}[t]
\begin{tcolorbox}[colback=gray!5, colframe=black, 
colbacktitle=gray!30, coltitle=black, title=Original Japanese, fonttitle=\bfseries, rounded corners]
与えられる画像と、そこに書かれている文字情報を用いて、\\
質問: [質問内容]\\
回答: [回答内容]\\
質問: [質問内容]\\
回答: [回答内容]\\
...\\
の形式でVQA問題を5問作ってください。解釈が曖昧になる主観的な問題ではなく、書かれている事実に基づいて客観的に判断できる問題を作ってください。またOCRのような文字の読み取り問題にはせず、内容理解を問う問題を作ってください。\\
画像内の文字:\\
\{OCR ANNOTATION HERE\}
\end{tcolorbox}

\begin{tcolorbox}[colback=gray!5, colframe=black, 
colbacktitle=gray!30, coltitle=black, title=Translated, fonttitle=\bfseries, rounded corners]
Using the given image and the textual information written in it, create 5 VQA questions in the following format:\\
Question: [Question content]\\
Answer: [Answer content]\\
Question: [Question content]\\
Answer: [Answer content]\\
...\\
Avoid subjective questions that could lead to ambiguous interpretations, and instead create questions that can be objectively answered based on the facts presented in the image. Also, do not include OCR-style text recognition questions; instead, create questions that test understanding of the image content. \\
Text in the image: \\
\{OCR ANNOTATION HERE\}
\end{tcolorbox}
\vspace{-1em}
\caption{Prompt for the synthetic VQA generation.}
\label{sup:prompt-genVQA}
\end{table}

\PAR{Prompt for Training and Evaluation}
For training and inference, we use task-specific prompts.
For the MangaOCR benchmark, we provide the prompt ``Please perform OCR on this image and output the recognized Japanese text along with its position (grounding)'' along with the input image. During training, the corresponding OCR annotations are included as supervision.
When running OCR inference with models other than the Qwen2.5 VL series, the outputs varied in format unless explicitly specified.
Therefore, we use the prompt in \Cref{sup:prompt-OCRinfer} to align their outputs with the OCR format used in the training data of MangaOCR.

\begin{table*}[t]
\begin{tcolorbox}[colback=gray!5, colframe=black, rounded corners]
Please perform OCR on this image and output the recognized Japanese text along with its position (grounding).\\ 
\\
The output should be a JSON list. Each item in the list must follow the structure below: \\
\textbackslash n\{"bbox\_2d": [x1, y1, x2, y2], "text\_content": "..."\} \\
\\
The field `"bbox\_2d"' must be a 2D bounding box that tightly encloses the text.  \\
Use the format `[x1, y1, x2, y2]', where: \\
- `x1', `y1' are the coordinates of the top-left corner of the bounding box, and \\
- `x2', `y2' are the coordinates of the bottom-right corner. \\
\\
Here is an example of the desired format: \\
\textbackslash n\{"bbox\_2d": [1490, 138, 1546, 201], "text\_content": "春休みです-"\} \\
\\
Please follow this format strictly.
\end{tcolorbox}
\vspace{-1em}
\caption{OCR inference prompt for models other than the Qwen2.5 VL series.}
\label{sup:prompt-OCRinfer}
\vspace{-2mm}
\end{table*}

For the MangaVQA benchmark, we use the prompt ``あなたは日本語の漫画に関する質問に答えるAIです。与えられた画像に基づいて質問に答えてください。(You are an AI that answers questions about Japanese manga. Please answer the given question based on the provided image.)'' together with the input image and a question. The ground-truth answer is given only during training.
For MangaVQA evaluation, the prompt in \Cref{sup:prompt-VQAeval} is used for LLM-as-a-judge.

\begin{table*}[t]
\begin{tcolorbox}[colback=gray!5, colframe=black, 
colbacktitle=gray!30, coltitle=black, title=System message, fonttitle=\bfseries, rounded corners]
You are an evaluator. Your task is to rate how appropriate a model's response is to a question about a manga image. For each case, you will be given a question (based on a manga image), a human-written answer, and the model's response. The image is not shown, but the question and answer are based on it. Please evaluate as if the image were available. \\
Please rate how well the model's response answers the question, considering the intended image context and the human answer as a reference, using a scale from 1 to 10: \\
1 — Completely inappropriate or unrelated to the question or image context. \\
2 — Mostly unrelated with major misunderstandings or incorrect information. \\ 
3 — Slightly relevant, but largely incorrect or unhelpful. \\
4 — Somewhat relevant, but contains significant errors or omissions. \\
5 — Partially correct with noticeable inaccuracies, vagueness, or missing key points. \\
6 — Generally okay, but missing core points or includes some incorrect interpretations. \\
7 — Mostly correct and relevant, with only minor issues or small omissions. \\ 
8 — Almost entirely accurate with only slight room for improvement. \\ 
9 — Very appropriate, accurate, and well-aligned with the question and image context. \\
10 — Perfectly appropriate, accurate, and fully answers the question as if the image were visible. \\
Only return a single number (1–10). Do not include any explanations, justifications, or comments. 
\end{tcolorbox}
\begin{tcolorbox}[colback=gray!5, colframe=black, 
colbacktitle=gray!30, coltitle=black, title=User prompt, fonttitle=\bfseries, rounded corners]
Input:\\
\hspace*{5mm}"question": \{question\},\\
\hspace*{5mm}"human-written answer": \{answer\},\\
\hspace*{5mm}"model's response": \{generated\_answer\},\\
Your score:
\end{tcolorbox}
\vspace{-1em}
\caption{Prompt for MangaVQA evaluation.}
\label{sup:prompt-VQAeval}
\end{table*}

\clearpage
\clearpage

\section{Additional Results}\label{sup-sec:additional-result}
We provide additional analysis and experimental results on our benchmarks, MangaVQA and MangaOCR.

\subsection{Effect of Model and Dataset Size} \label{subsec:model-data-size}
\Cref{tab:model-size} shows the performance of Qwen2.5-VL models of different sizes (3B and 7B) under various finetuning settings. 
Similar to the 7B model, the 3B model shows a slight drop in MangaOCR performance when finetuned on both $\mathrm{T_{OCR}}$ and $\mathrm{T_{VQA}}$, while its MangaVQA performance improves slightly.
\Cref{tab:data-size} shows the results of varying dataset size (25\%, 50\%, 75\%, and 100\%). 
We observe that performance generally improves as the dataset size increases.

\begin{table}[t]
    \centering
    \label{tab:model-size}
    \begin{adjustbox}{width=\linewidth}
    \begin{tabular}[t]{@{}c|l|c|c@{}}
        \toprule
         &  & \textbf{MangaOCR} & \textbf{MangaVQA} \\ 
        \textbf{Size}  & \textbf{FT data} & Hmean (\%) & LLM (/10.0) \\
        \midrule
        \multirow{4}{*}{3B} & None & 0.1 & 4.66 \\
         & $\mathrm{T_{OCR}}$ & 73.5 & 1.97 \\
         & $\mathrm{T_{VQA}}$ & 0.0 & 5.77 \\
         & $\mathrm{T_{OCR}}$+$\mathrm{T_{VQA}}$ & 66.5 & 6.01 \\
        \midrule
        \multirow{4}{*}{7B} & None & 0.9 & 5.65 \\
         & $\mathrm{T_{OCR}}$ & \textbf{74.9} & 1.03 \\
         & $\mathrm{T_{VQA}}$ & 0.0 & 6.54 \\
         & $\mathrm{T_{OCR}}$+$\mathrm{T_{VQA}}$ & 71.5 & \textbf{6.68} \\
        \bottomrule
    \end{tabular}
    \end{adjustbox}
    \vspace{-2mm}
    \caption{
    Effect of model size (3B and 7B).
    }
\end{table}

\begin{table}[t]
    \centering
    \label{tab:data-size}
    \begin{tabular}[t]{@{}l|c|c@{}}
        \toprule
         & \textbf{MangaOCR} & \textbf{MangaVQA} \\ 
        \textbf{Ratio} (\%) & Hmean (\%) & LLM (/10.0) \\
        \midrule
        25 & 59.0 & 6.20 \\
        50 & 64.9 & 6.20 \\
        75 & 68.4 & 6.48 \\
        100 & 71.5 & 6.68 \\
        \bottomrule
    \end{tabular}
    \vspace{-2mm}
    \caption{
    Effect of dataset size.
    }
\end{table}

\subsection{More Analysis of MangaVQA}

\paragraph{Comparison with Human Evaluation.}

\begin{figure}[t]
    \centering
    \includegraphics[width=\linewidth]{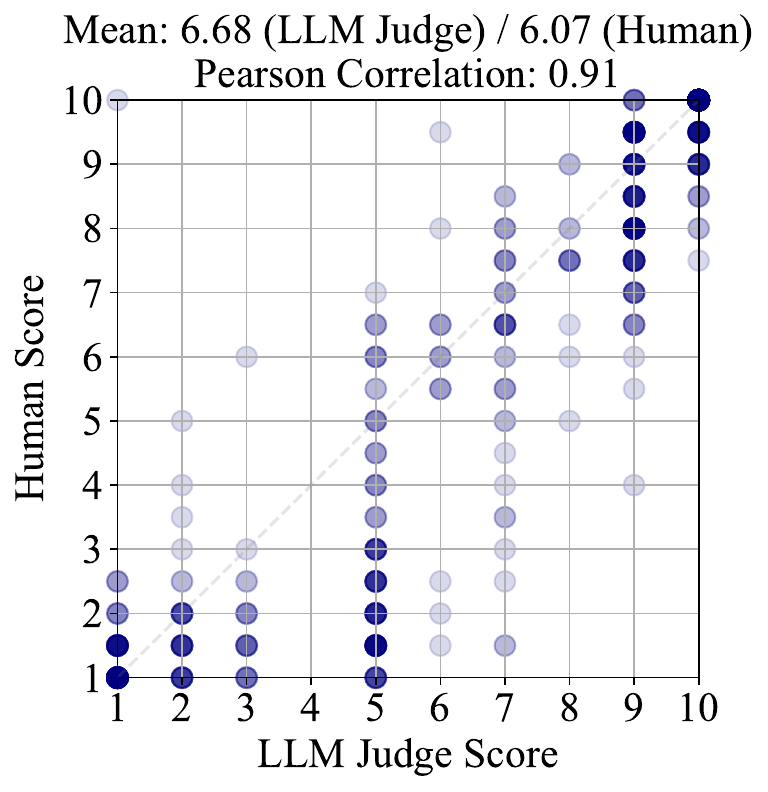}
    \vspace{-2em}
    \caption{
        \textbf{Comparison between LLM-Judge (Gemini) and Human Evaluation.} 
        Darker points indicate a higher concentration of points.
    }
    \label{fig:human_corr}
\end{figure}

To validate the reliability and consistency of the Gemini-judge employed in the MangaVQA evaluation, we conducted a comparative analysis between its evaluation scores and those provided by human annotators. Specifically, we asked two human evaluators to assign scores to all items in the benchmark dataset, following the same evaluation prompt used for the Gemini-judge.

The results of this comparison are illustrated in~\Cref{fig:human_corr}. We observe a small absolute difference in average scores ($\Delta = 0.61$). Additionally, there is a strong positive correlation between the scores assigned by the Gemini-judge and the human average ($r = 0.91$). These findings suggest that LLM-based evaluation can serve as a practical and consistent alternative to human judgment in our MangaVQA benchmark.

\PAR{Comparison of Results Using GPT-4o as the Judge}
Table~\ref{sup-tab:judge-result} compares the results when Gemini 2.5 Flash and GPT-4o are used as the LLM judge. 
The overall trends remain consistent across both settings. 
Here, a potential circular bias may exist when the same LLM is used for both generating responses and judging them.
In our case, the impact of such bias appears to be relatively minor.
Specifically, the performance difference of Gemini 2.5 Flash between being judged by itself and by GPT-4o is only 0.12.

\begin{table}[t]
    \tabcolsep=0.13cm
    \centering
    \begin{adjustbox}{width=\linewidth}
     \begin{tabular}[t]{@{}l|c|c@{}}
        \toprule
        \textbf{Method} & \textbf{Judge: Gemini} & \textbf{Judge: GPT-4o} \\ 
        \midrule
        GPT-4o & 6.00 & 5.76 \\
        Gemini 2.5 Flash & \textbf{7.26} & \textbf{7.14} \\
        Claude Sonnet 4.5 & 5.84 & 4.77 \\
        \midrule
        Phi-4-Multimodal-5.6B & 3.39 & 3.08 \\
        Pangea-7B & 3.23 & 2.98 \\
        LLaVA-OV-1.5-8B & 3.46 & 3.15 \\
        Sarashina2-Vision-8B & 4.45 & 4.13 \\
        Gemma-3-12B & 4.13 & 3.47 \\
        Heron-NVILA-Lite-15B & 3.76 & 3.32 \\ 
        Qwen2.5-VL 7B & 5.65 & 5.36 \\ 
        \midrule
        MangaLMM (Ours) & 6.68 & 6.57 \\
        \bottomrule
    \end{tabular}
    \end{adjustbox}
    \vspace{-2mm}
    \caption{
    \textbf{Comparison of LLM judge for MangaVQA.}
    }
    \label{sup-tab:judge-result}
\end{table}

\begin{table}[t]
    \centering
     \begin{tabular}[t]{@{}c|c@{}}
        \toprule
        \textbf{OCR Annot.} & \textbf{LLM (/10.0)} \\
        \midrule
        None & 5.64 \\
        Text & \textbf{6.68} \\
        Text + Pos. & 6.46 \\
        \bottomrule
    \end{tabular}
    \vspace{-2mm}
    \caption{
    Effect of OCR Annotation on VQA Generation.
    }
    \label{sup-tab:genVQA-OCRGT}
\end{table}

\begin{figure*}[t]
  \centering
  \includegraphics[width=\linewidth]{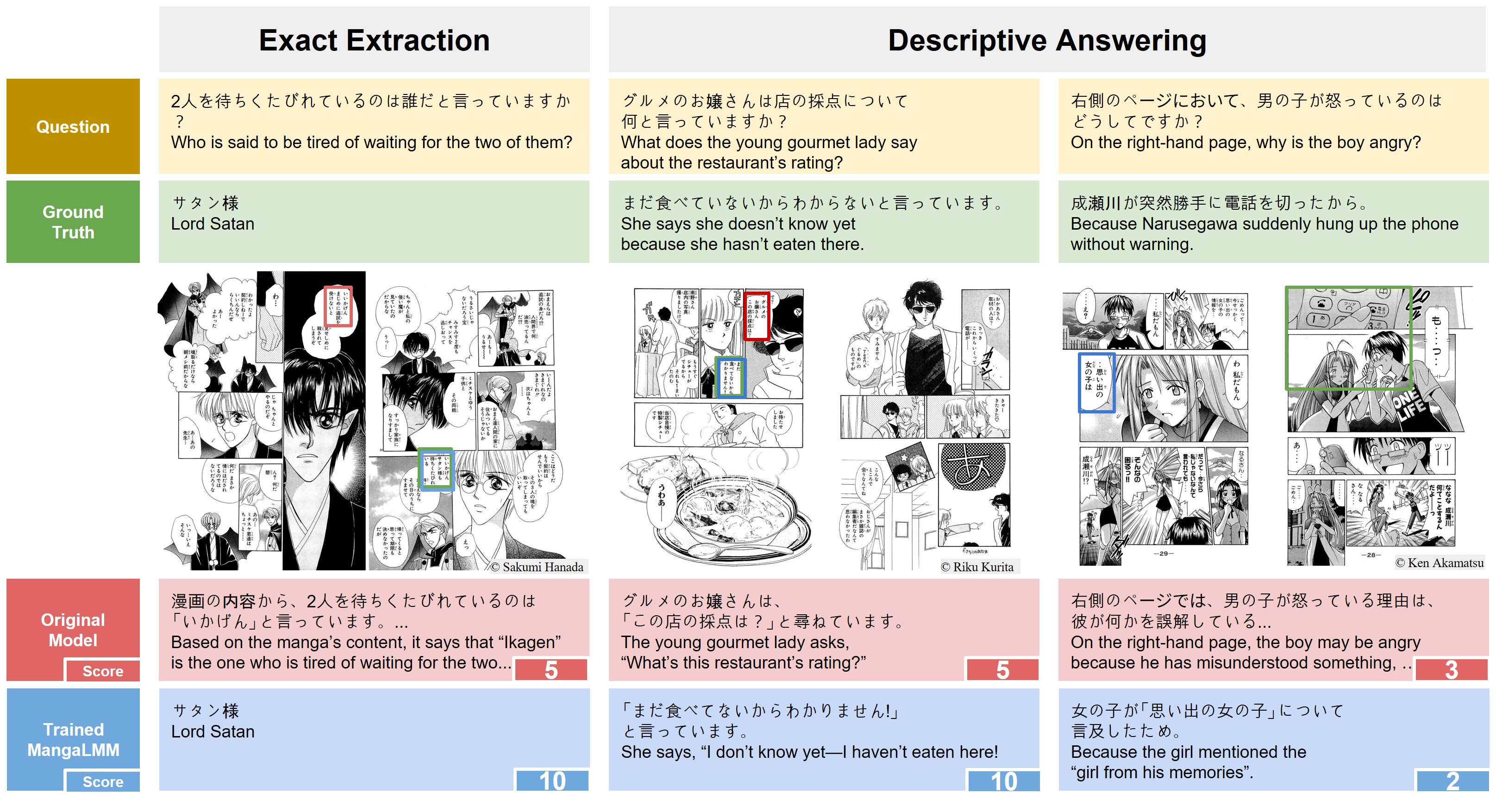}
  \vspace{-8mm}
  \caption{
    \textbf{Category-wise analysis on MangaVQA.}
    The regions in the image relevant to the question or models' answer are highlighted with boxes in corresponding colors.
  }
  \label{fig:quali_MangaVQA_More}
\end{figure*}

\PAR{More Analysis of OCR Annotation when Generating VQA Data} 
As described in \S\ref{subsec:Analysis-MangaVQA}, OCR annotation plays a key role in generating high-quality QA pairs with GPT-4o.
Here, we provide a more detailed analysis of the effect of OCR annotation. 
OCR annotation consists of both bounding box positions and their text content.
We compare the synthetic VQA data generated by GPT-4o using only the text content with those generated using both bounding box positions and text content.
Table~\ref{sup-tab:genVQA-OCRGT} presents the results.
Interestingly, our experiments show that using only the text content is more effective than including both text and positional information.
Although our current approach did not benefit from positional information, leveraging it remains a promising direction for future work.
Therefore, in our experiments, we use synthetic VQA examples generated using only the OCR text content.

\PAR{Qualitative Analysis of MangaVQA}
Figure \ref{fig:quali_MangaVQA_More} presents category-wise examples on MangaVQA.
For the categories on the left (Exact Extraction) and in the center (Descriptive Answering), the base Qwen 2.5-VL model often fails to locate the correct region and consequently extracts the wrong words. 
After finetuning, these issues are significantly improved in MangaLMM.
On the other hand, for the category on the right (Descriptive Answering), MangaLMM tends to over-prioritise text extraction, leading to incorrect answers even after training.

\begin{figure*}[t]
  \centering
  \begin{subfigure}{0.95\linewidth}
    \centering
    \includegraphics[width=\linewidth]{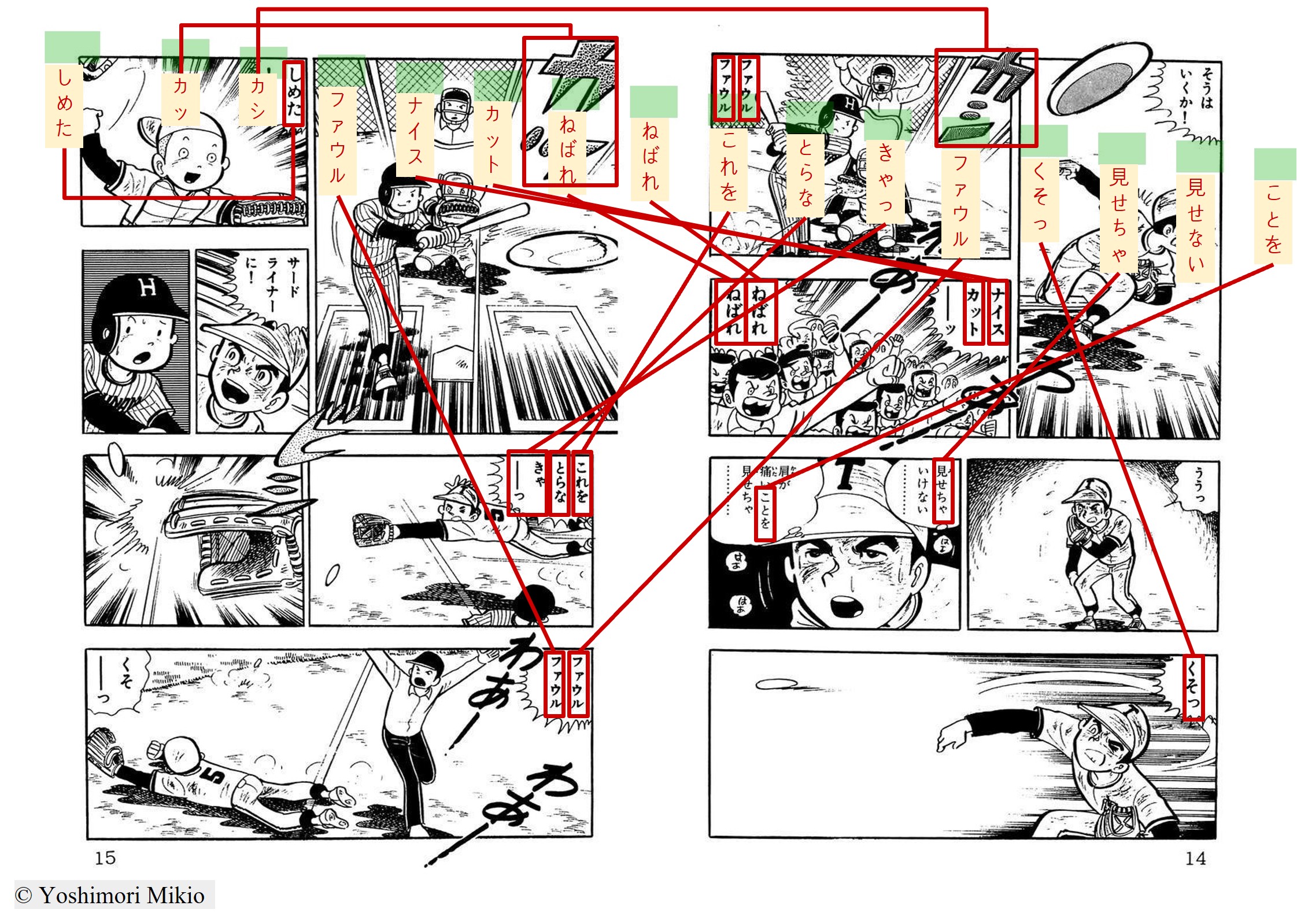}
    \vspace{-8mm}
    \caption{An example from the manga titled ShimatteIkouze.}
    \label{sup-fig:GPT-4o-OCR-ShimatteIkouze}
  \end{subfigure}
  
  \vspace{4mm}
  
  \begin{subfigure}{0.95\linewidth}
    \centering
    \includegraphics[width=\linewidth]{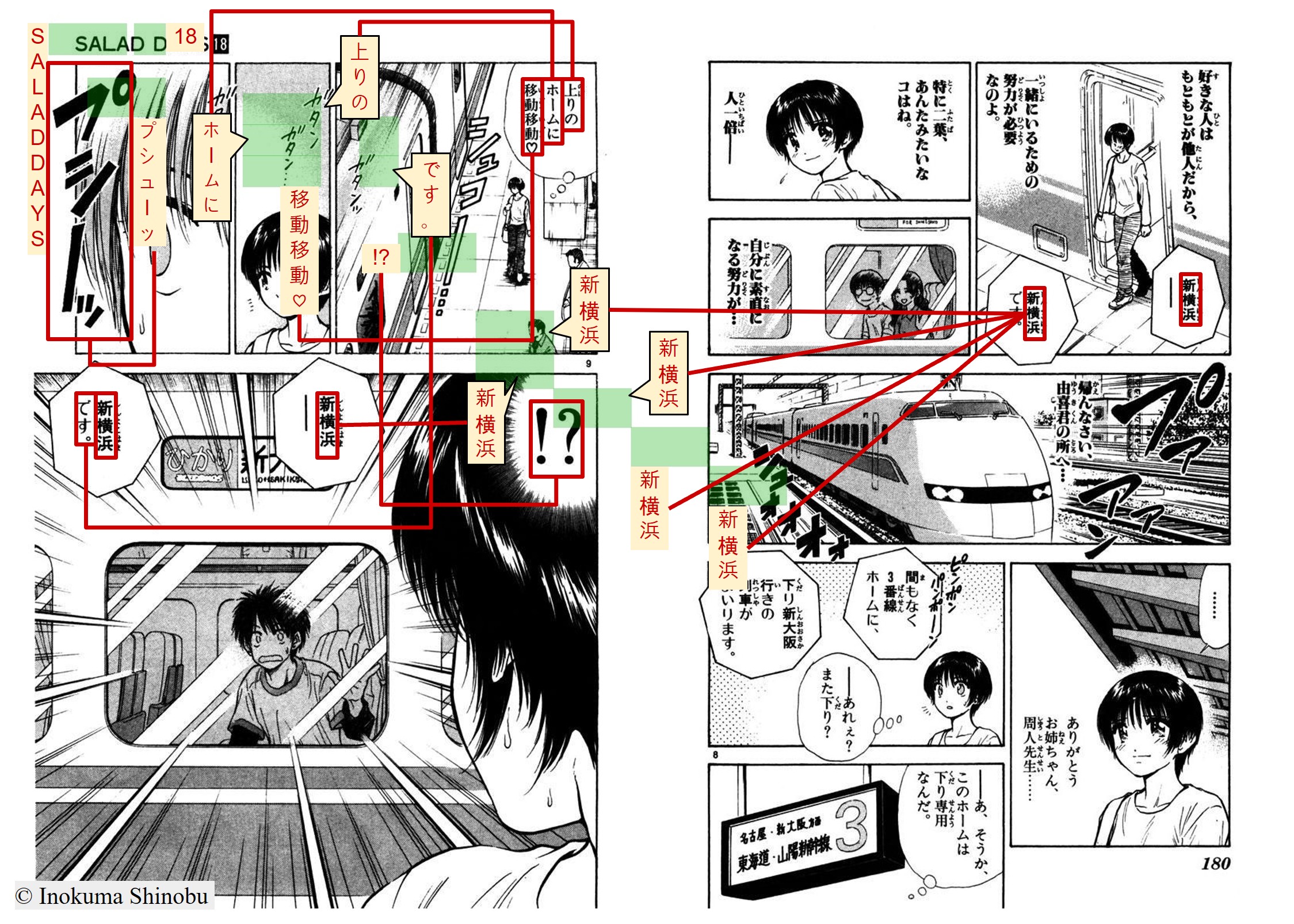}
    \vspace{-8mm}
    \caption{An example from the manga titled SaladDays.}
    \label{sup-fig:GPT-4o-OCR-SaladDays}
  \end{subfigure}
  \vspace{-3mm}
  \caption{
  \textbf{GPT-4o's Results on MangaOCR.}
  The green boxes indicate the detected text regions. The red text, shown near each green box, represents the predicted text fragment corresponding to that detected region.
  Each red bounding box is manually drawn to indicate where the predicted text fragment appears in the image.
  Red lines connect each predicted fragment to its corresponding detected position. These detected positions are almost always incorrect.
  }
  \label{sup-fig:GPT-4o-OCR}
\end{figure*}

\subsection{More Analysis of MangaOCR}\label{sup-sec:analysis-mangaocr}
We present a qualitative analysis of MangaOCR results from GPT-4o and MangaLMM.
As described in \S\ref{sec:experiment}, text segments that appear more than ten times are considered noise and excluded from the results. 
Therefore, such repeated segments do not appear in the visualizations.

\PAR{GPT-4o's Results on MangaOCR}\label{sup-sec:gpt4o-OCR}
Since previous studies have rarely conducted in-depth qualitative analysis of GPT-4o's OCR results, it is difficult to assess the model's actual performance on manga datasets.
We address this gap by providing a detailed qualitative analysis of GPT-4o's MangaOCR outputs.
\Cref{sup-fig:GPT-4o-OCR} shows GPT-4o's results on MangaOCR.
These examples demonstrate the low zero-shot OCR performance of GPT-4o in the manga domain.
The detected text regions almost always appear in incorrect or nonsensical locations, although the model can still read certain parts of the text within the image.
Because the predicted text positions are inaccurate, the outputs are considered entirely incorrect under OCR evaluation criteria.
While some predicted text fragments correspond to actual text in the image, there are many cases—such as in \Cref{sup-fig:GPT-4o-OCR}(b)—where most of the text is not recognized at all.
Even when text is recognized, it is often incorrect.
While GPT-4o fails to correctly detect and recognize most of the text, it can still recognize partial text content, which may allow GPT-4o to answer some text-based VQA questions.

Interestingly, when performing OCR inference with GPT-4o, the model sometimes generates disclaimers such as:
``The bounding box coordinates and text content are illustrative and may not perfectly match the actual image. For precise OCR and bounding box extraction, specialized OCR tools like Tesseract or Google Vision API should be used.''
This suggests that GPT-4o itself acknowledges its limitations in precise OCR and recommends using dedicated OCR tools.

\begin{figure*}[t]
  \centering
  \begin{subfigure}{0.95\linewidth}
    \centering
    \includegraphics[width=\linewidth]{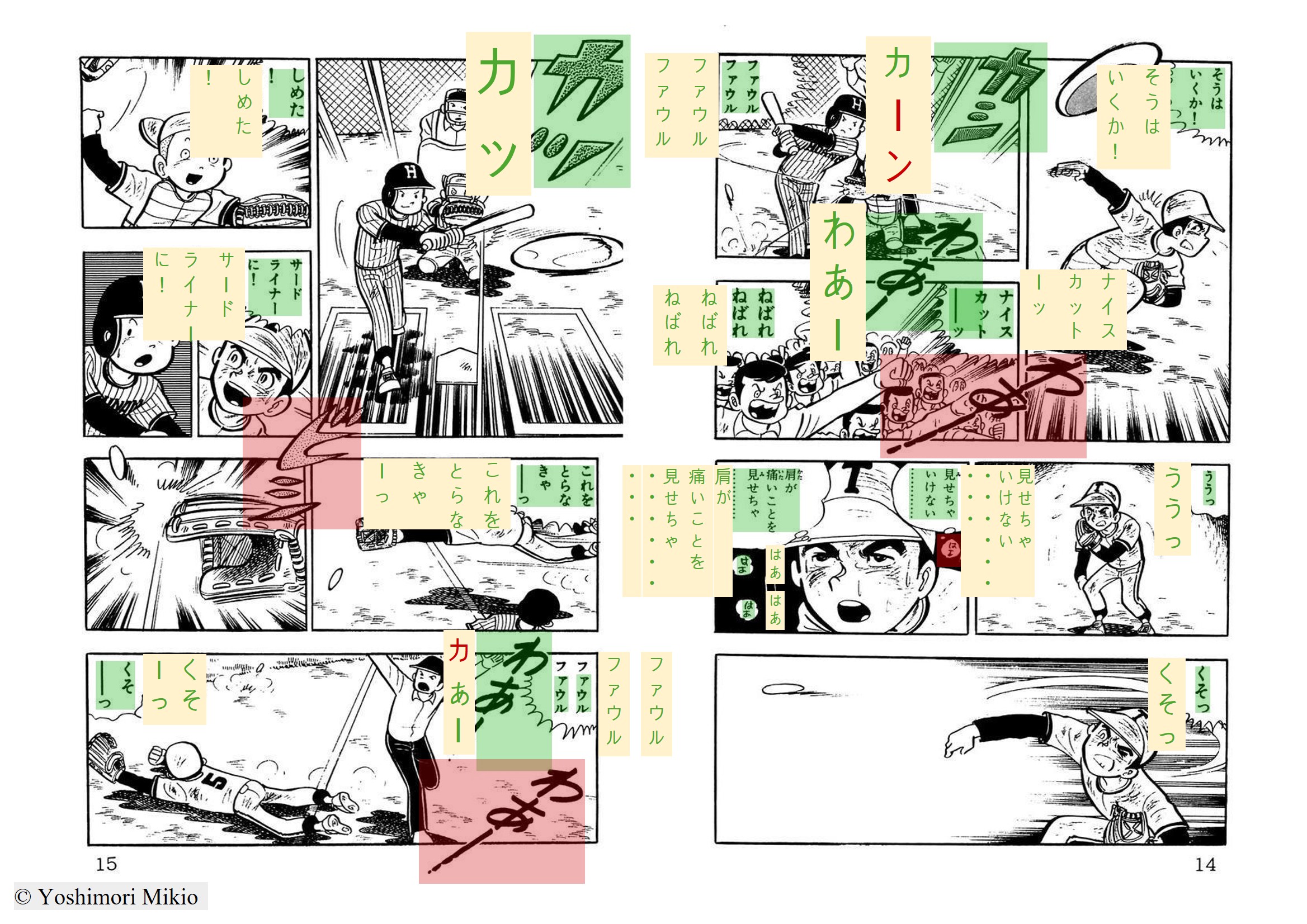}
    \vspace{-8mm}
    \caption{An example from the manga titled ShimatteIkouze.}
    \label{sup-fig:MangaLMM-OCR-ShimatteIkouze}
  \end{subfigure}
  
  \vspace{4mm}
  
  \begin{subfigure}{0.95\linewidth}
    \centering
    \includegraphics[width=\linewidth]{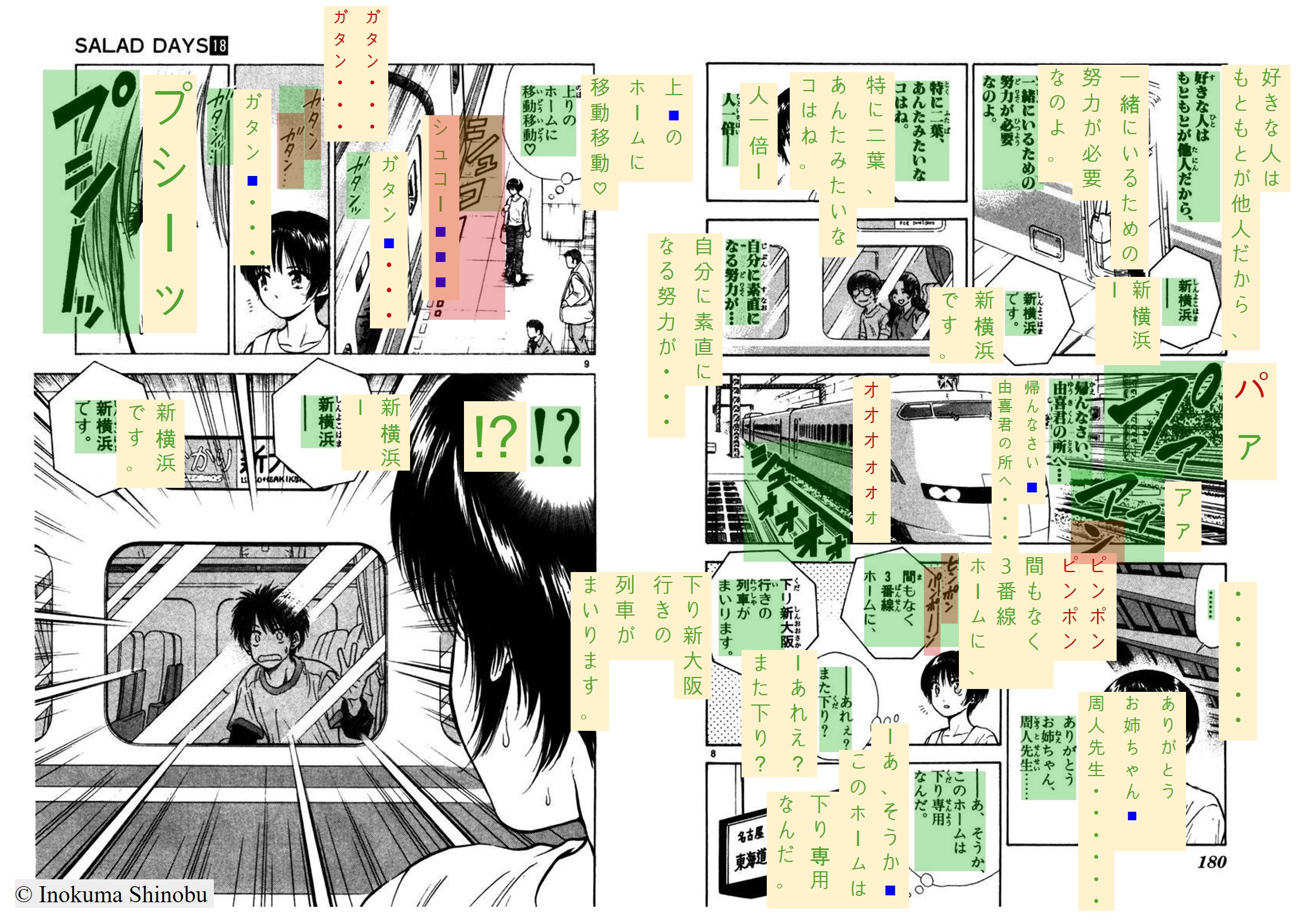}
    \vspace{-8mm}
    \caption{An example from the manga titled SaladDays.}
    \label{sup-fig:MangaLMM-OCR-SaladDays}
  \end{subfigure}
  \vspace{-3mm}
  \caption{
    \textbf{MangaLMM's Results on MangaOCR.}
    The green boxes indicate the detected text regions. 
    The text shown near each green box is the predicted text for that detected region.
    The green text represents correctly predicted text, while the red text indicates incorrectly predicted text.
    Missing characters are marked with small blue squares.
    The red boxes show false negatives—text regions that should be detected but are missed.
    Most OCR results are correct.
  }
  \label{sup-fig:MangaLMM-ocr}
\end{figure*}

\PAR{MangaLMM's Results on MangaOCR}\label{sup-sec:quali-OCR}
\Cref{sup-fig:MangaLMM-ocr} shows MangaLMM's results on MangaOCR.  
As seen in the figure, most predictions appear correct, reflecting the model's strong OCR capability across a wide range of text sizes, from large to small.  
The red regions indicate false negatives. Occasionally, even text that appears large and seemingly easy to detect is missed.  
According to our manual inspection, such cases are mostly onomatopoeia.  
This suggests that the model struggles more with onomatopoeic expressions, which are often written in non-standard fonts, sizes, or orientations, compared to regular text.

\PAR{MangaOCR Evaluation without Positional Information}
What if MangaOCR were evaluated without considering positional information? 
To further analyze the models' text recognition ability, we evaluate them under a setting that does not depend on positional information.
TextMonkey~\cite{liu2024textmonkey} represents the first study to perform OCR evaluation without positional information, and several follow-up works (e.g., CC-OCR~\cite{yang2024cc}) have adopted its evaluation approach.
Following the evaluation method and code provided by TextMonkey, we adopt the so-called ``Trans'' mode, which ignores positional alignment and instead checks whether each ground-truth string appears anywhere within the predicted text.

In this evaluation, all predicted text strings from the image are concatenated into a single sequence, and each ground-truth instance is evaluated by checking whether its text appears within the predicted string. 
Each ground-truth instance must be exactly and completely included in the prediction, and even a one-character mismatch results in an incorrect outcome. 
This stricter criterion generally yields lower scores than the edit distance–based evaluation, which assigns partial scores even when some characters in the text string are incorrect.
For example, if the ground truth is ``apple'' and the concatenated prediction is ``banana apple orange,'' the instance is considered correct. 
However, if the concatenated prediction is ``banana aple orange,'' it is counted as incorrect due to the missing character.

The results are summarized in Table~\ref{sup-tab:OCR-Textmonkey}.
Under this setting, several models such as GPT-4o, Gemini 2.5 Flash, Claude Sonnet 4.5, Gemma-3, and Qwen 2.5-VL achieved non-zero scores. 
However, models without finetuning still exhibited consistently low performance. 
This highlights the importance of finetuning on OCR-specific data.

\begin{table}[t]
    \tabcolsep=0.13cm
    \centering
    \begin{adjustbox}{width=\linewidth}
     \begin{tabular}[t]{@{}l|c|c@{}}
        \toprule
        \textbf{Method} & \textbf{Traditional OCR metric} & \textbf{TextMonkey-Trans} \\ 
        \midrule
        GPT-4o & 0.0 & 12.9 \\
        Gemini 2.5 Flash & 0.0 & 11.7 \\
        Claude Sonnet 4.5 & 0.0 & 7.3 \\
        \midrule
        Phi-4-Multimodal-5.6B & 0.0 & 0.0 \\
        Pangea-7B & 0.0 & 0.0 \\
        LLaVA-OV-1.5-8B & 0.0 & 0.0 \\
        Sarashina2-Vision-8B & 0.0 & 0.7 \\
        Gemma-3-12B & 0.0 & 3.7 \\
        Heron-NVILA-Lite-15B & 0.0 & 0.0 \\ 
        Qwen2.5-VL 7B & 0.9 & 5.7 \\ 
        \midrule
        MangaLMM (Ours) & 71.5 & 63.2 \\
        \bottomrule
    \end{tabular}
    \end{adjustbox}
    \vspace{-2mm}
    \caption{
    Model performance under two evaluation settings: the Traditional OCR metric and the TextMonkey ``Trans'' mode (ignoring positional information).
    }
    \label{sup-tab:OCR-Textmonkey}
\end{table}

\PAR{Comparison with dedicated OCR system.}
We evaluated DeepSolo++~\cite{ye2023deepsolo++}, a multilingual OCR system, on the MangaOCR task. The model achieved an Hmean of 5.4\%, indicating very low performance in this domain. Since DeepSolo++ operates zero-shot on manga-style images, we observed that it often detects only small fragments of the text (e.g., one or two characters) rather than identifying the full text in the speech balloons, leading to many incorrect predictions. This behavior is consistent with what we observe in other LMMs when applied zero-shot to manga.

\clearpage

\begin{table}[t]
    \tabcolsep=0.13cm
    \centering
    \begin{adjustbox}{width=\linewidth}
     \begin{tabular}[t]{@{}l|c|c|c|c@{}}
        \toprule
        & \textbf{MangaOCR} & \textbf{MangaVQA} & \textbf{MMMU} & \textbf{MMBench} \\ 
        \textbf{Method} & Hmean (\%) & LLM (/10.0) & Acc. (\%) & Acc. (\%) \\
        \midrule
        GPT-4o & 0.0 & 6.00 & 70.7 & 89.0 \\
        Gemini 2.5 Flash & 0.0 & 7.26 & 79.6 & - \\
        \midrule
        Phi-4-Multimodal-5.6B & 0.0 & 3.39 & 55.1 & 86.7 \\
        Qwen2.5-VL 7B & 0.9 & 5.65 & 58.6 & 87.8 \\
        \midrule
        MangaLMM & 71.5 & 6.68 & 25.8 & 1.5 \\
        \bottomrule
    \end{tabular}
    \end{adjustbox}
    \vspace{-2mm}
    \caption{
    Comparison of LMMs on MangaOCR, MangaVQA, MMMU, and MMBench.
    }
    \label{sup-tab:manga+natural}
\end{table}

\begin{table}[t]
    \tabcolsep=0.13cm
    \centering
    \begin{adjustbox}{width=\linewidth}
     \begin{tabular}[t]{@{}l|c|c|c|c@{}}
        \toprule
        & \textbf{MangaOCR} & \textbf{MangaVQA} & \textbf{MMMU} & \textbf{MMBench} \\ 
        \textbf{FT data} & Hmean (\%) & LLM (/10.0) & Acc. (\%) & Acc. (\%) \\
        \midrule
        $\mathrm{T_{OCR}}$+$\mathrm{T_{VQA}}$ (MangaLMM)  & 71.5 & 6.68 & 25.8 & 1.5  \\
        $\mathrm{T_{OCR}}$+$\mathrm{T_{VQA}}$ + LO-50K & 70.2 &  6.56 &  49.6 & 82.0 \\
        \bottomrule
    \end{tabular}
    \end{adjustbox}
    \vspace{-2mm}
    \caption{
    Finetuned results on a combined dataset including natural image understanding data.
    LO-50K denotes LLaVA Onevision 50K data. 
    }
    \label{sup-tab:finetune}
\end{table}

\subsection{Exploring the Capability for Manga and Natural Image Understanding}
To further investigate the capability of LMMs across both manga and natural images, we conducted additional evaluations and experiments. 

\paragraph{Evaluation on natural image benchmarks.}
We evaluated general-purpose baselines (GPT-4o, Gemini 2.5 Flash, Phi-4-Multimodal, and Qwen2.5-VL-7B) as well as our MangaLMM on two representative benchmarks commonly used for assessing LMM performance on natural images: MMMU~\cite{yue2024mmmu} and MMBench~\cite{liu2024mmbench}. 
The table below (an extended version of Table~\ref{tab:main-results}) shows the results.
Scores for models other than MangaLMM are obtained from the official MMMU leaderboard or the Phi-4 technical report~\cite{phi-4MM}.
As expected, the general-domain performance of MangaLMM (finetuned from Qwen2.5-VL-7B) drops noticeably due to domain specialization, underscoring the challenge of maintaining broad visual understanding after task-specific finetuning.

\paragraph{Joint finetuning with natural image data.}
We also finetuned our model on a combined dataset consisting of manga data and 50K randomly sampled natural image examples from the LLaVA-OneVision dataset~\cite{li2024llava}. 
As shown in the table below (an extended version of Table~\ref{tab:FT}), this joint finetuning substantially restores the model's performance on natural image benchmarks, while preserving strong performance on our MangaOCR and MangaVQA benchmarks.

These results demonstrate that although domain specialization may reduce general capability, it can be effectively recovered through joint training—highlighting the feasibility of developing LMMs that are both manga-capable and broadly applicable to natural image understanding.

\begin{table}[t]
    \tabcolsep=0.13cm
    \centering
    \begin{adjustbox}{width=\linewidth}
     \begin{tabular}[t]{@{}l|c|c|c@{}}
        \toprule
        \textbf{Method} & \textbf{Total (100 images)} & \textbf{JPN subset (6 images)	} & \textbf{MangaOCR} \\ 
        \midrule
        Qwen2.5-VL 7B & 2.8 & 0.3 & 0.9 \\
        MangaLMM & 13.5 & 50.0 & 71.5\\
        \bottomrule
    \end{tabular}
    \end{adjustbox}
    \vspace{-2mm}
    \caption{OCR evaluation on other comics from the eBDtheque dataset and MangaOCR.
    }
    \label{sup-tab:eBDtheque}
\end{table}

\PAR{Evaluation on Other Comics}
We find that MangaLMM shows some degree of generalization to other comics, although its performance is naturally lower than on in-domain Japanese manga. As discussed in the related work, there is currently no established VQA dataset for the comic domain, making it difficult to use a standard VQA benchmark in this setting. To examine cross-domain performance, we therefore conducted an additional OCR evaluation on the well-known comic dataset eBDtheque~\cite{eBDtheque2013}, which contains 100 images from European, American, and Japanese comics. 
Table~\ref{sup-tab:eBDtheque} shows the results.

The performance of MangaLMM on eBDtheque is lower than on MangaOCR, reflecting the domain differences between the datasets. Notably, MangaLMM still outperforms Qwen2.5-VL, suggesting that training on manga contributes positively to OCR performance even in other comic styles. We attribute the lower performance on eBDtheque to two main factors: (1) Page format mismatch: MangaOCR consists of two-page spreads, whereas eBDtheque images are single-page. This mismatch frequently leads to shifted or misaligned predicted bounding boxes. (2) Color vs. grayscale: MangaOCR is entirely black-and-white, while more than half of eBDtheque images are in color. MangaLMM, trained on grayscale manga, shows degraded performance on colored pages. We expect both issues to be addressable by incorporating single-page layouts and color comic images during training. While MangaLMM is intentionally specialized for manga, extending it toward a general comic-capable LMM is a promising direction for future work.

\end{document}